\pdfoutput=1

\documentclass[11pt]{article}

\usepackage{acl}

\usepackage{times}
\usepackage{latexsym}

\usepackage[T1]{fontenc}

\usepackage[utf8]{inputenc}

\usepackage{microtype}

\usepackage{inconsolata}

\usepackage{graphicx}

\usepackage{amsmath}
\usepackage{multirow}
\usepackage{amssymb}
\usepackage{float}
\usepackage{longtable}
\usepackage{tabularx}
\newcommand{\subsectiontitle}[1]{\noindent\textbf{#1}\hspace{0.5em}}
\usepackage{pifont} 
\usepackage{amssymb}
\usepackage{booktabs} 
\usepackage{enumitem} 
\usepackage{colortbl} 
\usepackage{xcolor} 
\usepackage{array} 
\usepackage{makecell} 
\usepackage{tabularray} 
\usepackage{arydshln} 
\usepackage{colortbl}  
\usepackage{xspace}
\newcommand{\ours}{\textsc{LIFBench}\xspace}
\newcommand{\ourseval}{\textsc{LIFEval}\xspace}
\usepackage[most]{tcolorbox} 
\usepackage{algorithm,  algorithmicx}      
\usepackage{algpseudocode} 

\floatname{algorithm}{Program}
\tcbset{
promptbox/.style={
enhanced,
colback=white, 
colframe=black, 
colbacktitle=gray!20, 
coltitle=black,
fonttitle=\bfseries, 
fontupper=\normalsize, 
width=\linewidth, 
boxrule=0.5mm, 
arc=3mm, 
left=1mm, right=1mm, top=1mm, bottom=1mm, 
boxsep=5pt, 
rounded corners, 
}}
\title{\ours: Evaluating the Instruction Following Performance and Stability of Large Language Models in Long-Context Scenarios
}

\author{%
Xiaodong~Wu$^{\dagger}$ 
\quad Minhao~Wang$^{\dagger}$ 
\quad Yichen~Liu $^{\dagger}$ 
\textbf{\quad Xiaoming Shi $^{\dagger}$ } \\
\textbf{\quad He~Yan $^{\S}$ }
\textbf{\quad Xiangju~Lu $^{\S}$} 
\textbf{\quad Junmin~Zhu $^{\S}$} \textbf{\quad Wei~Zhang} $^{\dagger\ddagger}$\thanks{Corresponding author.} \\
$^{\dagger}$East China Normal University
\quad
$^{\S}$iQIYI Inc 
\quad
$^{\ddagger}$Shanghai Innovation Institute
\\
  \tt\small 
  $^{\dagger}$\{51255901079,51275901104,51275901148\}@stu.ecnu.edu.cn 
  \quad xmshi@ir.hit.edu.cn \\
  \tt\small
  \quad $^{\S}$\{yanhe, luxiangju, zhujunmin\}@qiyi.com \quad \textsuperscript{*}zhangwei.thu2011@gmail.com
}

\begin{document}
\maketitle
\begin{abstract}
As Large Language Models (LLMs) evolve in natural language processing (NLP), their ability to stably follow instructions in long-context inputs has become critical for real-world applications. 
However, existing benchmarks seldom focus on instruction-following in long-context scenarios or stability on different inputs.
To bridge this gap, we introduce \ours, a scalable dataset designed to evaluate LLMs' instruction-following capabilities and stability across long contexts. \ours comprises three long-context scenarios and eleven diverse tasks, featuring 2,766 instructions generated through an automated expansion method across three dimensions: length, expression, and variables.
For evaluation, we propose \ourseval, a rubric-based assessment method that enables precise, automated scoring of complex LLM responses without reliance on LLM-assisted assessments or human judgment. 
This method allows for a comprehensive analysis of model performance and stability from multiple perspectives.
We conduct detailed experiments on 20 prominent LLMs across six length intervals.
Our work contributes \ours and \ourseval as robust tools for assessing LLM performance in complex and long-context settings, offering valuable insights to guide future advancements in LLM development.
\footnote{Data and code are available at \url{https://github.com/SheldonWu0327/LIF-Bench-2024}}

\end{abstract}

\section{Introduction}

\begin{table}[t]
  \centering
  \renewcommand{\arraystretch}{0.95}
  \resizebox{\columnwidth}{!}{
  \begin{tabular}{lcccc}
  \arrayrulecolor{black}
    \Xhline{1pt}
    \rowcolor[gray]{0.9}
    \textbf{Benchmark} & \textbf{Long.} & \textbf{Inst.} & \textbf{Stab.}  & \textbf{Unlim.}  \\
    \hline
    ZeroSCROLLS (\citeyear{shaham2023zeroscrolls})  & \textcolor{blue}{\ding{51}}  & \textcolor{red}{\ding{55}}   & \textcolor{red}{\ding{55}} & \textcolor{red}{\ding{55}}  \\
    BAMBOO (\citeyear{dong2024bamboo})              & \textcolor{blue}{\ding{51}}   & \textcolor{red}{\ding{55}}   & \textcolor{red}{\ding{55}} & \textcolor{red}{\ding{55}}  \\
    Longbench (\citeyear{bai2023longbench})         & \textcolor{blue}{\ding{51}}   & \textcolor{red}{\ding{55}}   & \textcolor{red}{\ding{55}} & \textcolor{red}{\ding{55}}  \\
    $\infty$ Bench (\citeyear{zhang2024infinitebench})     & \textcolor{blue}{\ding{51}}   & \textcolor{red}{\ding{55}}   & \textcolor{red}{\ding{55}} & \textcolor{red}{\ding{55}}  \\
    RULER (\citeyear{hsieh2024ruler})               & \textcolor{blue}{\ding{51}}   & \textcolor{red}{\ding{55}}   & \textcolor{red}{\ding{55}} & \textcolor{blue}{\ding{51}}  \\ 
    \hdashline
    IFEval (\citeyear{ifezhou2023instruction})      & \textcolor{red}{\ding{55}}   & \textcolor{blue}{\ding{51}}   & \textcolor{red}{\ding{55}} & \textcolor{red}{\ding{55}}  \\ 
    FollowBench (\citeyear{jiang2023followbench})   & \textcolor{red}{\ding{55}}   & \textcolor{blue}{\ding{51}}   & \textcolor{red}{\ding{55}} & \textcolor{red}{\ding{55}}  \\ 
    InfoBench (\citeyear{qin2024infobench})         & \textcolor{red}{\ding{55}}   & \textcolor{blue}{\ding{51}}   & \textcolor{red}{\ding{55}} & \textcolor{red}{\ding{55}}  \\ 
    CELLO (\citeyear{he2024can})                    & \textcolor{red}{\ding{55}}   & \textcolor{blue}{\ding{51}}   & \textcolor{red}{\ding{55}} & \textcolor{red}{\ding{55}}  \\
    \hline
    \ours (Ours)                                 & \textcolor{blue}{\ding{51}}   & \textcolor{blue}{\ding{51}}   & \textcolor{blue}{\ding{51}} & \textcolor{blue}{\ding{51}}  \\
    \Xhline{1pt}
  \end{tabular}}
  \vspace{-1.5ex}
  \caption{A comparison of our \ours with some relevant datasets. We summarize their focus, including \underline{Long}-context scenarios, \underline{Inst}ruction-following, and model \underline{Stab}ility. 'Unlim.' denotes whether the data length can be \underline{Unlim}ited. 
  }
  \label{tab:comparison}
  \vspace{-2ex}
\end{table}

\begin{figure*}
    \centering
    \includegraphics[width=1\linewidth]{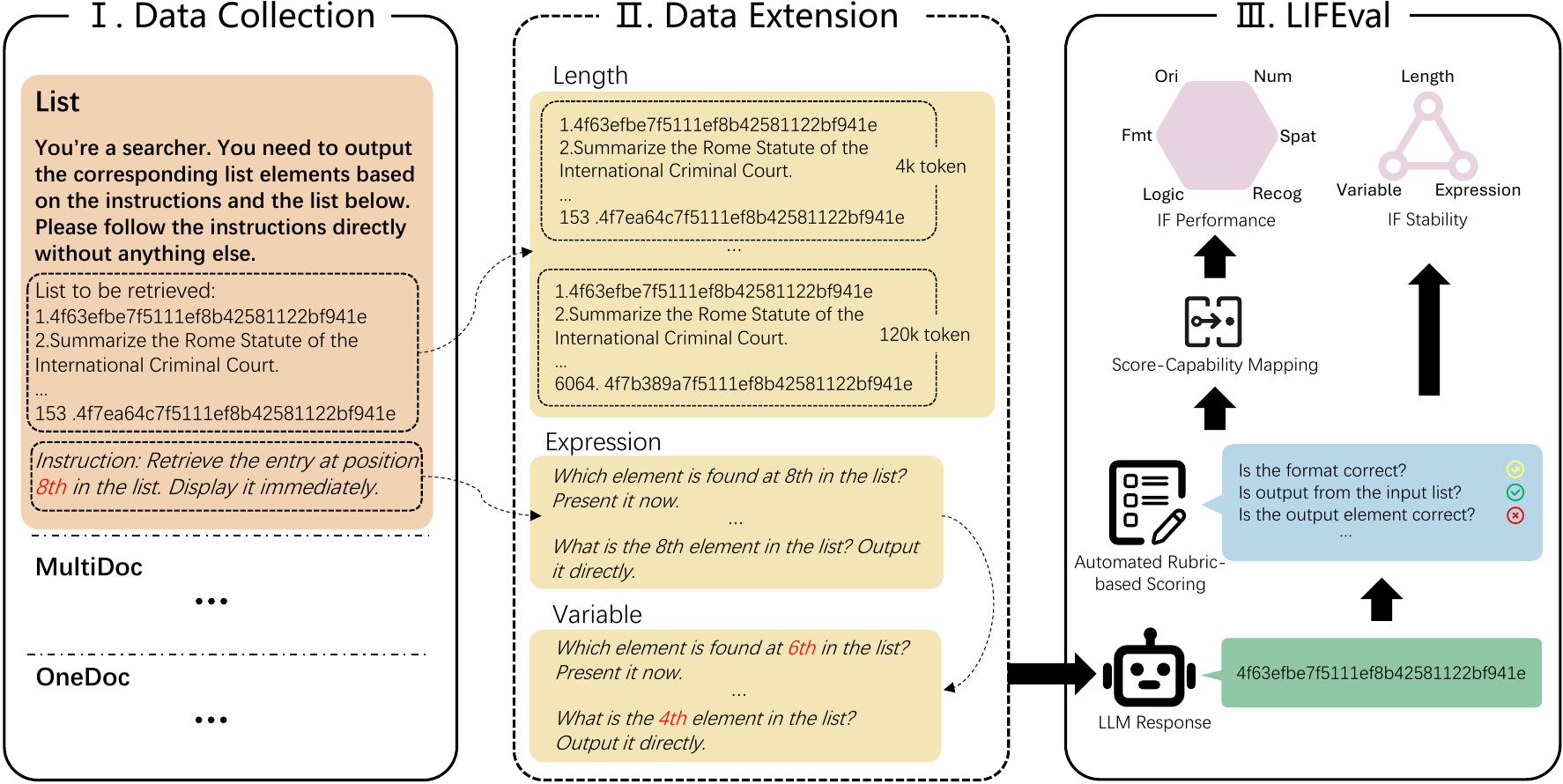}
    \vspace{-4ex}
    \caption{The framework of \ours, where the task \textit{Single-ID} in the List scenario is used as an example. \textbf{Bold} denotes the scenario description $D$; normal denotes the context $X$; \textit{italics} denotes instruction $I$, with \textcolor{red}{red} indicate the instruction variables $var$, and the remaining black parts correspond to the instruction template $tpl$.}
    \label{fig:overview}
    \vspace{-3.3ex}
\end{figure*}
As Large Language Models (LLMs) continue to make significant strides across practical applications~\cite{achiam2023gpt, chowdhery2023palm, brown2020language}, their performance in natural language processing (NLP) tasks has reached unprecedented levels.
These tasks span text generation~\cite{que2024hellobench, tan2024proxyqa, zhang2024benchmarking}, complex reasoning~\cite{parmar2024logicbench,DBLP:journals/corr/abs-2503-13109}, and problem-solving~\cite{lu2023mathvista,li2024gsm}.
Despite these achievements, significant challenges remain. 
On the one hand, LLMs often struggle to accurately and consistently follow human instructions, such as restating input content precisely or adhering stably to specific formatting constraints~\cite{he2024can}. On the other hand, studies show that as input length increases, the LLMs' performance in tasks such as reasoning~\cite{levy2024same}, retrieval~\cite{li2024needlebench}, and general NLP~\cite{bai2023longbench} deteriorates. 
These challenges pose substantial barriers to their effectiveness in real-world applications.

Numerous evaluation benchmarks, as summarized in Table~\ref{tab:comparison}, have been proposed to guide the development of LLMs. However, they each exhibit notable limitations when it comes to evaluating instruction-following capabilities and stability in long-context scenarios.
Some benchmarks focus either on long-context scenarios~\cite{bai2023longbench, zhang2024infinitebench} or complex instruction-following abilities~\cite{ifezhou2023instruction, jiang2023followbench}. However, none of these works evaluate instruction-following abilities in long-context scenarios, and their reliance on fixed data lengths fails to accommodate the state-of-the-art LLMs' ever-expanding context length.
Other efforts~\cite{li2024needlebench, hsieh2024ruler} attempt to extend evaluation to longer contexts by constructing synthetic datasets, but their tasks are hard to provide a comprehensive and in-depth assessment of LLMs' instruction-following abilities.
In terms of evaluation outcomes, most existing benchmarks focus exclusively on task completion performance, often neglecting stability—a critical factor in ensuring reliable real-world performance.

To address these limitations, we introduce the \textbf{L}ong-context \textbf{I}nstruction \textbf{F}ollowing \textbf{Bench}mark (\ours), a scalable benchmark for evaluating LLMs' instruction-following capability and stability in long-context scenarios. The framework of our benchmark is shown in Figure \ref{fig:overview}. 
Considering real-world scenarios, we construct three long-context scenarios based on the granularity of information to be processed. 
On this basis, eleven delicate tasks are designed, which can illustrate various dimensions of instruction-following capabilities. 
We manually craft templates for all tasks, and introduce an automated instruction expansion method from three dimensions (length, expression, and variables), enabling \ours to expand significantly in both the quantity and length of instructions. As an example, we construct a dataset of 2,766 instructions spanning six length intervals, reaching up to 128k tokens. 

For evaluation, traditional metrics for downstream tasks are often unsuitable for complex instruction-following scenarios \cite{honovich2023unnatural}. Moreover, while many studies rely on GPT-4 for automated and open-ended assessment, these approaches encounter limitations due to notable gaps between GPT-4 and human experts~\cite{qin2024infobench, jiang2023followbench}, as well as potential bias problems~\cite{wang2023largebias}.
To address these challenges, we propose \ourseval, a systematic and efficient evaluation method for complex LLM responses, without relying on LLMs or human evaluators. 
Specifically, by designing task-specific scoring rubrics, we decompose evaluations into fine-grained and quantifiable scoring points, each of which can be assessed automatically. 
In addition, through the score-capability map and a novel metric—IFS, \ourseval provides insights into models' fundamental capabilities and stability from various perspectives.

Overall, our contributions are as follows:
\begin{itemize}[nosep, left=0pt, itemsep=1pt, topsep=1pt]
    \item We introduce \ours, a benchmark designed to evaluate instruction-following capabilities in long-context scenarios, containing 11 tasks across three scenarios.
    \item We develop methods for dataset expansion across three perspectives, enabling high scalability in both the quantity and length of instructions. 
    \item We propose \ourseval, an automatic evaluation method for accurately and comprehensively assessing the quality and stability of LLMs' complex responses.
    \item We conduct extensive experiments across six length intervals, which evaluate and analyze the instruction-following capabilities and stability of 20 well-known LLMs, encompassing both open-source and closed-source models.
\end{itemize}

\section{Related Work}

Several studies focus on LLMs' performance in long contexts. These benchmarks collect data from traditional NLP tasks (e.g., summarization and question answering (QA)) to form comprehensive datasets containing long data~\cite{ shaham2022scrolls,shaham2023zeroscrolls,dong2024bamboo, li2024long,Gavin2024LongInsAC}.
Additionally, due to the excellent performance of LLMs on open-ended tasks, some benchmarks have also designed synthetic tasks to better probe the models' ability in math, reasoning, and logic~\cite{kwan2023m4le,zhang2024infinitebench, wang2024leave, li2024needlebench,DBLP:conf/emnlp/ChenCZH024}.
 For evaluation, some works adopt regular metrics (e.g., Acc, ROUGE, and BLEU), which can be obtained by automated calculations. However, some open-ended tasks cannot be effectively evaluated using these metrics, hence powerful LLMs, such as GPT-4, are used as alternative evaluators. For example, the studies~\cite{an2023leval,li2023loogle} feed the model predictions and ground-truth answers to the GPT-4 evaluator, which is tasked with conducting a direct scoring or comparison to evaluate baselines' performance on partial tasks. Unlike \ours, these benchmarks only assess problem-solving capabilities in long-context scenarios, overlooking challenges of complex instruction following that arise in real-world applications.

Given the complexities of evaluating instruction-following abilities, several studies introduce meaningful innovations in their evaluation methodologies to better tackle this multifaceted challenge.
The studies~\cite{cook2024ticking, wen2024complexbench,qin2024infobench,zhang2024cfbench} decompose instructions into checklists composed of YES/NO questions or PASS/FAIL criteria, which answers should meet. CELLO~\cite{he2024can} defines four criteria that should be assessed as they can encompass common errors made by LLMs, and develops automated evaluation metrics to reflect models' ability in complex instruction-following scenarios. 
However, these studies fail to concern evaluation in long-context instruction-following scenarios and largely overlook stability in the instruction-following process. 
Additionally, while the study~\cite{Sakai2024TowardTE} explores the impact of various prompt templates and languages on LLM, it focuses solely on NLU tasks.

In summary, existing benchmarks either emphasize long-context scenarios or instruction-following capabilities, whereas  \ours uniquely targets both simultaneously. 

\section{\ours}
\subsection{Problem Definition}
As shown in Figure~\ref{fig:overview}, we model the instruction-following task in long-context scenarios as follows: Given a prompt consisting of a scenario description ($D$), context ($X$), and instruction ($I$), the model is expected to output an answer ($A$). This process can be represented as:
\begin{equation}
  \label{eq:modeling}
  (D,X_{len},I_{tpl, var}) \to A \,.
\end{equation}

In this setup, the scenario description $D$ provides task background at the beginning of the prompt, and all tasks within a scenario share the same $D$. 
The context $X$, as the main body of the prompt, provides essential information and varies by scenario. For example, in the List scenario (three scenarios are constructed in 
\ours, see Section \ref{Datacollection}), $X$ is a long list, while in the OneDoc scenario, $X$ represents a lengthy processed document. The parameter $len$ represents the number of tokens in $X$. 
The instruction $I$ is placed at the end of the prompt, which consists of two components: (1) the instruction template ($tpl$), outlining the task requirements, and (2) the instruction variable ($var$), representing variable part within the template. 
Generally, in \ours, $D$ and $I$ tend to be short, while $X$ is a long text with thousands of tokens.

\subsection{Dataset Construction}
\label{Datacollection}

\begin{figure}[t]
    \centering
    \includegraphics[width=0.86\linewidth]{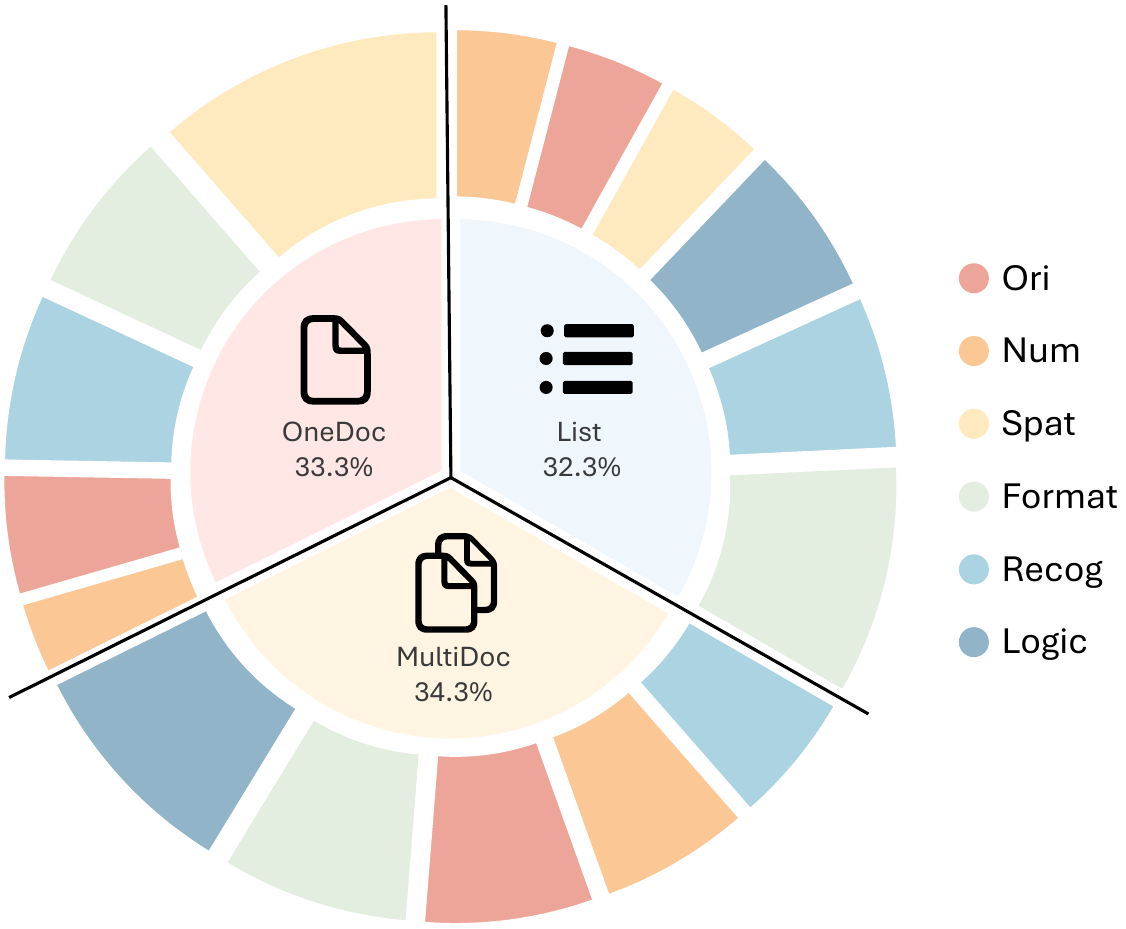}
    \vspace{-1ex}
    \caption{Rubric score distribution across different scenarios and capabilities. 
    Scores across different scenarios and capabilities tend to be evenly distributed.
    }
    \vspace{-1ex}
    \label{fig:stastic}
    \vspace{-1ex}
\end{figure}

To simulate real-world LLM applications in long-text processing, we construct 3 scenarios and 11 tasks (see Table \ref{fig:stastic}) based on the following principles:
(1) \textbf{Task Diversity}: Tasks should encompass varied constraints (e.g., format, quantity) to evaluate different instruction-following abilities; (2) \textbf{Performance Distinguishability}: Tasks must balance simplicity and complexity to distinguish model performance; (3) \textbf{Input Scalability}: Tasks should support extended input lengths to assess long-context capabilities; (4) \textbf{Automated Evaluation}: Task constraints can be assessed through an automated program.
\vspace{0.5ex}

\subsectiontitle{List}
The List scenario tests how well LLMs can handle structured lists, such as retrieving specific items and processing structured data. The input $X$ is an ordered list with UUIDs and natural language instructions.

The scenario includes six tasks: \textit{Single-ID} and \textit{Multi-ID} focus on retrieving specific elements from an ordered list using provided IDs. Building on these basic tasks, the \textit{Offset-ID}, \textit{Offset-element}, \textit{Blur-ID}, and \textit{Blur-element} tasks introduce spatial constraints, adding complexity to the retrieval process. "Offset" tasks require precise index-based retrieval to test fine-grained spatial awareness, while "Blur" tasks involve broader spatial ranges, allowing more flexibility. "ID" and "Element" refer to the type of reference in the retrieval process, representing either the position number in the ordered list or the list element itself.

\subsectiontitle{MultiDoc}
The MultiDoc scenario evaluates how well LLMs process multiple documents, such as summarization and retrieval. Models need to compare documents, find differences, and handle batch operations.

The input $X$ consists of multi-document collections from diverse sources. Each document has six fields: "text", "id", "iD2", "title", "date", and "source", with a length of 300–500 tokens. Tasks include \textit{Find-dup-doc}, which finds duplicate documents, and \textit{Batch-label}, which assigns labels to documents based on given attributes. 

\subsectiontitle{OneDoc}
The OneDoc scenario tests how well LLMs process a single long document, such as extracting key information or answering questions.

A long document is created by combining essays from the Paul Graham Essays dataset\footnote{\url{https://huggingface.co/datasets/sgoel9/paul_graham_essays}}. Some sentences are marked as key information. Tasks include \textit{Repeat}, where models repeat a given amount of key information; \textit{Extract}, where models extract specific key details; and \textit{QA}, where models check if a sentence is labeled as key information.

\vspace{1ex}
We manually write scenario descriptions and instruction templates for all tasks, with detailed examples provided in Appendix \ref{sec:exampleinLIFBench}. To further ensure the tasks and scenarios are challenging and discriminative, special efforts are made during data collection, as elaborated in Appendix \ref{appdix:datacollection}.
\subsection{Data Extension}
\label{sec: Data Extension}
In this section, we expand manual templates in three dimensions (length, expression, and variable) to form a sizeable test dataset.

\subsectiontitle{Length} 
A number of works \cite{bai2023longbench, ni2024xl, li2024long, levy2024same} have found that the length of input text has an impact on the ability of the LLMs. In \ours, different lengths of prompts allow us to explore the impact of context length on the instruction-following capabilities of LLMs, making it essential to introduce variations in prompt length. 

In all three scenarios, we adjust the length of the prompt by controlling the context token count $l$. Specifically, we modify the number of elements in the List or the number of documents in MultiDoc and OneDoc to achieve the desired length. Ample corpus are pre-constructed for each scenario, supporting expansions up to 2M tokens in one prompt.

\subsectiontitle{Expression}
In real-world contexts, due to personality and individual differences, individuals often provide significantly varied descriptions of the same subject, which undoubtedly challenges the stability of large models in following instructions. 
To assess LLM robustness in this regard, we diversify instruction templates to create multiple expressions style with differing wording and syntax.

Our approach follows a four-step process, namely "Rewriting-Encoding-Clustering-Sampling (RECS)".
First, to ensure diversity and mitigate biases from any single model, we use GPT-4 \cite{achiam2023gpt} and Claude \cite{anthropic2024claude} to generate 40+ rewrites of each original instruction template, complemented by a subsequent manual review to further validate and refine the outputs. Next, the rewritten templates are encoded into vector representations for clustering, with the number of clusters set to the target number of rewritten instructions for each task. For instance, if a task needs five rewritten instructions, we will create five clusters. Finally, from each cluster, we select the usable template nearest to the center as the final diversified expression. Further details and effectiveness evidence are provided in Appendix~\ref{appdix: effective EE}.

\begin{table}[t]
  \centering
  \resizebox{.99\columnwidth}{!}{
  \begin{tabular}{llcccc}
  \rowcolor[gray]{0.9}
    \Xhline{1pt}
    \textbf{Scenario} & \textbf{Task} & \textbf{ID} & \textbf{\#Exp.} & \textbf{ \#Var.} &  \textbf{\#Data} \\
    \hline
    \multirow{6}{*}{\underline{L}ist}   & \underline{S}ingle-\underline{I}D         & LSI    & 5 & 6 &  180 \\
                            & \underline{M}ulti-\underline{I}D          & LMI  & 5 & 5 &  150 \\
                            & \underline{O}ffset-\underline{I}D         & LOI  & 11 & 6  &  396 \\
                            & \underline{O}ffset-\underline{E}lement    & LOE  & 12 & 6  &  432 \\
                            & \underline{B}lur-\underline{I}D           & LBI  & 11 & 6  &  396 \\
                            & \underline{B}lur-\underline{E}lement      & LBE  & 12 & 6  &  432 \\ \hline
    \multirow{2}{*}{\underline{M}ultiDoc}    
                            & \underline{B}atch-label      & MB    & 5 & 5 &  150 \\
                            & \underline{F}ind-dup-doc     & MF    & 5 & 5 &  150 \\ \hline
    \multirow{3}{*}{\underline{O}neDoc} 
                            & \underline{R}epeat            & OR    & 5 & 5 &  150 \\
                            & \underline{Q}A                & OQ    & 5 & 6 &  180 \\ 
                            & \underline{E}xtract           & OE    & 5 & 5 &  150 \\
    \Xhline{1pt}
  \end{tabular}}
 \vspace{-1ex}
  \caption{Statistics of \ours. \textbf{\#Exp.} and \textbf{\#Var.} represent the count in the data extension of \underline{Exp}ression and \underline{Var}iable.}
  \vspace{-4ex}
  \label{tab:stastic-info}

\end{table}

\subsectiontitle{Variable}
Some placeholders are preset in the instruction templates to indicate variable parts of instructions, i.e., instruction variables. These variables encompass elements such as query keys (for retrieval tasks), categorization criteria (for classification tasks), and format requirements. For position-related or numerical variables, we maintain an even distribution, and manually adjust other variables during task iterations to precisely control the task difficulty (see examples in Appendix \ref{app:instruction-variable}).
By analyzing LLM performance across varying instruction variables, we can evaluate the models' understanding and consistency in executing instructions. 
Ideally, models with strong instruction-following abilities execute instructions stably across variable conditions, while weaker models may exhibit inconsistency. 

\vspace{1ex}
Table \ref{tab:stastic-info} details the task counts for each scenario. Compared to the other two scenarios, tasks in the List scenario are simpler, so they carry less weight (see Section \ref{sec:ars}) but are more numerous.

\vspace{-0.5ex}
\section{\ourseval}
\vspace{-0.5ex}
In this section, we introduce \ourseval, an automatic method that provides accurate assessments and detailed insights into LLMs' long-context instruction-following capabilities and stability.

\subsection{Automated Rubric-based Scoring} 
\label{sec:ars}
To evaluate the output quality of LLMs on \ours tasks, we introduce Automated Rubric-based Scoring (ARS), an accurate and efficient programmatic evaluation method. 
As shown in Table \ref{tab:rubric} in Appendix \ref{app:rubric design}, we manually craft scoring rubrics $\mathcal{R}$ for each task.
Each rubric consists of several scoring points $s$ and is assigned a weight $\tilde s$ according to its complexity and difficulty. 
In other words, a larger $\tilde s$ means greater complexity, requiring more steps for evaluation.
All of these points can be assessed automatically through a program. The scoring process on task $t$ can be shown as follows:
\begin{equation}
    \begin{split}
    ARS_{t}(\mathcal{A}_t)=\frac{1}{\tilde {R_t}}\sum_{s \in \mathcal{R}_t}f_s(\mathcal{A}_t),
    \tilde R_t=\sum_{s \in \mathcal{R}_t}\tilde s \,.
    \end{split}
    \label{eq:ars}
\end{equation}

In this equation, $\mathcal{A}_t$ represents the model's responses on task $t$, and $\tilde R_t$ is the sum of weights across all scoring points. The function $f_s(\cdot) \in [0, \tilde s]$ represents the average score of all outputs for scoring point $s$.
Its implementation relies on programmatic evaluation pipelines tailored to each scoring point. For example, in structural verification (e.g., JSON Dict format, see Program~\ref{alg:json}), we use a multi-stage detection process that includes symbol check, Parsing check, and KV Check. These stages progressively impose stricter requirements on the model's output, with final scores based on performance at each stage. 
Overall, while the core logic is manually engineered according to the rubric, the execution is fully automated through programs. We provide detailed considerations regarding the scoring rubric, as well as additional examples of evaluation programs, in Appendix \ref{app:rubric design}.

Naturally, defining $\mathcal{T}=\{t_i\}_{i=1}^{N_t}$ as the set of all tasks, the overall test result for any model on \ours is the weighted average of scores across~$\mathcal{T}$:
\begin{equation}
ARS_{overall}=\frac{\sum_{t \in \mathcal{T}}\tilde R_t\cdot ARS_t }{\sum_{t\in \mathcal{T}}\tilde R_t} \,.
\label{eq: overall ars}
\end{equation}

Although a well-designed scoring rubric enables \ourseval to provide a more comprehensive and rigorous evaluation, the relatively high human cost inevitably limits its generalizability.
To address this, when extending \ours with new samples or adapting \ourseval to other datasets, users may adopt a simplified approach by replacing weighted averaging in Equation \ref{eq:ars} and \ref{eq: overall ars} with a simple mean. This streamlined method bypasses the rubric design process while still satisfying the requirements of many practical applications.

\begin{table*}[t]
    \centering
    \small
    \resizebox{\textwidth}{!}{
    
    \begin{tabular}{lcccp{1pt}ccccccp{1pt}ccp{1pt}c}
     \arrayrulecolor{black}
        \Xhline{1pt}
        \rowcolor[gray]{0.9} 
         & \multicolumn{3}{c}{\textbf{OneDoc}} & & \multicolumn{6}{c}{\textbf{List}} & & \multicolumn{2}{c}{\textbf{MultiDoc}} & &\\
        \noalign{\vskip -2pt} \cmidrule(lr){2-4} \cmidrule(lr){6-11}  \cmidrule(lr){13-14}
         \rowcolor[gray]{0.9} \noalign{\vskip -3pt}
        \multirow{-2}{*}{\textbf{Model}}& \textbf{OR} & \textbf{OQ} & \textbf{OE} & & \textbf{LSI} & \textbf{LMI} & \textbf{LOI} & \textbf{LOE} & \textbf{LBI} & \textbf{LBE} & & \textbf{MB} & \textbf{MF} & &\multirow{-2}{*}{\textbf{Overall}}\\ \hline
        
        \multicolumn{16}{c}{\textit{API Models}} \\ \hdashline
            GPT-4o & \textbf{0.797} & \textbf{0.882} & \underline{0.834} & ~ & \underline{0.881} & \textbf{0.836} & \textbf{0.740} & \textbf{0.823} & 0.749 & \underline{0.825} & ~ & 0.719 & 0.588 & ~ & \textbf{0.758 } \\ 
        GPT-4 & 0.707 & 0.820 & 0.736 & ~ & \textbf{0.893} & 0.735 & \underline{0.704} & 0.781 & 0.750 & \textbf{0.832} & ~ & 0.600 & \textbf{0.777} & ~ & \underline{0.738}  \\ 
        
        \hline \multicolumn{16}{c}{\textit{Models Larger Than 20B Parameters}}\\ \hdashline
        Qwen2.5-72B-Inst.\textsuperscript{$\dagger$} & \underline{0.759} & 0.774 & 0.817 & ~ & 0.867 & 0.674 & 0.680 & 0.681 & \textbf{0.818} & 0.806 & ~ & 0.609 & 0.584 & ~ & 0.706  \\ 
        Llama-3.1-70B-Inst. & 0.730 & \underline{0.860} & 0.711 & ~ & 0.805 & \underline{0.798} & 0.651 & \underline{0.798} & 0.693 & 0.788 & ~ & 0.657 & 0.531 & ~ & 0.694  \\ 
        Qwen2.5-32B-Inst.\textsuperscript{$\dagger$} & 0.702 & 0.792 & \textbf{0.850} & ~ & 0.662 & 0.612 & 0.542 & 0.517 & \underline{0.763} & 0.761 & ~ & 0.608 & 0.476 & ~ & 0.650  \\ 
        C4AI-cmd-r-08-2024 (32B)\textsuperscript{$\dagger$} & 0.529 & 0.838 & 0.535 & ~ & 0.692 & 0.619 & 0.494 & 0.530 & 0.667 & 0.615 & ~ & \underline{0.729} & \underline{0.660} & ~ & 0.626  \\ 
        C4AI-cmd-r-v01 (35B)\textsuperscript{$\dagger$} & 0.495 & 0.818 & 0.420 & ~ & 0.641 & 0.579 & 0.489 & 0.506 & 0.694 & 0.643 & ~ & 0.721 & 0.646 & ~ & 0.595  \\ 
        Qwen2.5-72B\textsuperscript{$\dagger$} & 0.402 & 0.694 & 0.428 & ~ & 0.604 & 0.579 & 0.453 & 0.497 & 0.642 & 0.666 & ~ & 0.726 & 0.522 & ~ & 0.548  \\ 
        Llama-3.1-70B & 0.337 & 0.273 & 0.118 & ~ & 0.668 & 0.394 & 0.525 & 0.581 & 0.695 & 0.681 & ~ & 0.708 & 0.308 & ~ & 0.422  \\ 
        Qwen2.5-32B\textsuperscript{$\dagger$} & 0.347 & 0.529 & 0.280 & ~ & 0.263 & 0.334 & 0.232 & 0.306 & 0.338 & 0.297 & ~ & \textbf{0.732} & 0.380 & ~ & 0.394 \\ 
        
        \hline\multicolumn{16}{c}{\textit{Models With 7-20B Parameters}}\\ \hdashline
        Qwen2.5-14B-Inst.\textsuperscript{$\dagger$} & 0.593 & 0.768 & 0.637 & ~ & 0.525 & 0.601 & 0.457 & 0.385 & 0.700 & 0.570 & ~ & 0.591 & 0.349 & ~ & 0.547  \\ 
        InternLM2.5-7b-chat-1m & 0.446 & 0.828 & 0.378 & ~ & 0.609 & 0.438 & 0.543 & 0.631 & 0.713 & 0.764 & ~ & 0.619 & 0.428 & ~ & 0.523  \\ 
        Qwen2.5-7B-Inst.\textsuperscript{$\dagger$} & 0.507 & 0.812 & 0.447 & ~ & 0.626 & 0.568 & 0.436 & 0.531 & 0.684 & 0.701 & ~ & 0.445 & 0.436 & ~ & 0.519  \\ 
        Llama-3.1-8B-Inst. & 0.537 & 0.681 & 0.413 & ~ & 0.705 & 0.535 & 0.397 & 0.537 & 0.668 & 0.637 & ~ & 0.522 & 0.308 & ~ & 0.491  \\ 
        GLM-4-9b-chat-1m & 0.484 & 0.813 & 0.267 & ~ & 0.705 & 0.534 & 0.371 & 0.514 & 0.648 & 0.667 & ~ & 0.688 & 0.300 & ~ & 0.490  \\ 
        Qwen2.5-14B\textsuperscript{$\dagger$} & 0.273 & 0.550 & 0.257 & ~ & 0.339 & 0.326 & 0.281 & 0.290 & 0.359 & 0.290 & ~ & 0.697 & 0.397 & ~ & 0.384  \\ 
        LWM-Text-Chat-1M (7B) & 0.413 & 0.730 & 0.075 & ~ & 0.633 & 0.291 & 0.309 & 0.605 & 0.590 & 0.590 & ~ & 0.128 & 0.520 & ~ & 0.381  \\ 
        Llama-3.1-8B & 0.347 & 0.287 & 0.040 & ~ & 0.600 & 0.207 & 0.422 & 0.554 & 0.625 & 0.622 & ~ & 0.471 & 0.455 & ~ & 0.375  \\ 
        Qwen2.5-7B\textsuperscript{$\dagger$} & 0.268 & 0.491 & 0.233 & ~ & 0.106 & 0.113 & 0.068 & 0.102 & 0.119 & 0.149 & ~ & 0.244 & 0.233 & ~ & 0.213  \\ 
        LWM-Text-1M (7B) & 0.307 & 0.252 & 0.049 & ~ & 0.164 & 0.136 & 0.110 & 0.306 & 0.420 & 0.452 & ~ & 0.112 & 0.220 & ~ & 0.204 \\  
        \Xhline{1pt}
    \end{tabular}
    }
    \vspace{-1ex}
    \caption{The ARS scores of models on different tasks. The abbreviations of the tasks can be found in Table \ref{tab:stastic-info}. The overall score is calculated by Eq. \ref{eq: overall ars}. The best performing score is highlighted in \textbf{bold} and second-best is \underline{underlined}. $\dagger$ indicates that the context $X$ on the longest interval is right-truncated.}
    \vspace{-4ex}
    \label{tab:main results}

\end{table*}

\subsection{Score-Capability Mapping}
To further offer insights into the model's strengths and weaknesses across various dimensions of instruction following, we introduce Score-Capability Mapping, which maps the scoring point $s$ to six fundamental capabilities.
With reference to previous studies~\cite{he2024can, zhou2023instruction}, the six capabilities are defined as follows:

\subsectiontitle{Ori} (Original Content): Abilities to reproduce the original input accurately.

\subsectiontitle{Num} (Numerical Ability): Abilities in handling numerical data, such as recognition, counting, and basic arithmetic.

\subsectiontitle{Spat} (Spatial Awareness): Abilities in understanding spatial relationships and sequences.

\subsectiontitle{Fmt} (Format): Abilities in modifying and structuring content according to format rules.

\subsectiontitle{Logic} (Logic Execution): Abilities to follow logical conditions and decision branches.

\subsectiontitle{Recog} (Recognition Ability): Abilities to differentiate and focus on key elements of the input.


Building on the ARS score, we further compute the Instruction Following Performance (IFP) for each capability. The IFP for a specific capability \( c \) is defined as:
\begin{equation}
    \label{eq:ifp}
    IFP_c = \frac{\sum_{t\in \mathcal{T}}\sum_{s\in \mathcal{R}_t}\mathbb{I}(s,c)\cdot f_s(\mathcal{A})}{\sum_{t\in \mathcal{T}}\sum_{s\in \mathcal{R}_t}\mathbb{I}(s,c)\cdot \tilde s}\,
\end{equation}
where \( \mathbb{I}(s,c) \) is the indicator function that identifies whether a scoring point \( s \) is related to the capability \( c \). We manually crafted this mapping so that when a scoring point \( s \) is relevant to the capability \( c \), the indicator \( \mathbb{I}(s,c) \) equals 1, and 0 otherwise.
Table~\ref{tab:rubric} provides more detailed information on this.
 Importantly, as shown in Figure \ref{fig:stastic}, scores across different scenarios and capabilities are carefully balanced during data collection to ensure less bias in \ourseval.

\subsection{Instruction Following Stability}

To measure whether the model can consistently follow instructions, we introduce Instruction Following Stability (IFS).
Specifically, we define the observation perspective of stability, $p$, which refers to a specific feature of the model input. In \ours, there are three perspectives: prompt length, expression (i.e., the template of instruction~$I$), and instruction variables. Based on the selected perspective, we partition model's responses $\mathcal{A}_t$ into $N_p$ groups, denoted as $\mathcal{A}_t^p = \{ \mathcal{A}_t^j \}_{j=0}^{N_p}$. For example, a set of responses \(\mathcal{A}_t\) may originate from inputs spanning five different length intervals (such as \textit{4k}, \textit{8k}, \textit{etc.}). In this case, we group the responses into five distinct groups based on the input lengths. Subsequently, ARS scores for each group of answers will be computed, resulting in a set of performance values $\mathcal{Y}_t^p$, which can be expressed as follows:
\begin{equation}
    \begin{split}
        \mathcal{Y}_t^p = \{ ARS_t(\mathcal{A}_t^j) \}_{j=0}^{N_p}\,.
    \end{split}
\end{equation}
Finally, the Instruction Following Stability (IFS) on task $t$ is calculated as the standard deviation (denoted by $\sigma(\cdot)$) of the performance values $Y$ divided by their mean, formally expressed as:
\begin{equation}
    \label{eq:ifs}
    \begin{split}
        IFS_t^p = \frac{\sigma(\mathcal{Y}_t^p)}{\bar {\mathcal{Y}}_t^p} \in [0, +\infty)
    \end{split}
\end{equation}
A lower IFS indicates greater stability in instruction-following under perspective \( p \), whereas a higher IFS signifies reduced stability.



\section{Experiments}
\subsection{Experiment Setup}
We evaluate 20 popular LLMs with long-context capabilities, including models from GPT \cite{achiam2023gpt}, Llama \cite{dubey2024llama}, Qwen \cite{qwen2}, C4AI \cite{cohere_for_ai_2024}, LWM \cite{Liu2024WorldMO}, InternLM \cite{cai2024InternLM2}, and GLM \cite{glm2024chatglm} series, all claiming to support context lengths exceeding 128k tokens. Notably, the Qwen2.5-Inst. model extends its 32k context length to 128k with YaRN \cite{Peng2023YaRNEC}. GPT-4 and GPT-4o are accessed via its official API, while open-source models are deployed using vLLM \cite{kwon2023efficient}.

The experiments were conducted across six context length intervals, ranging from 4k to 128k tokens, with task-specific output limits to ensure sufficient space for model generation. 
Token counts are calculated using GPT-4's tokenizer\footnote{\url{https://github.com/openai/tiktoken}}, and truncation is applied to adjust context~$X$ for models unable to process the longest contexts. 
In data extension, 5–6 template variants are created for each original instruction, and each instruction variable has 5–10 candidates for sampling.
Further details can be found in Appendix \ref{app:experimentsetup}.

\begin{table}[t]
  \centering
  \resizebox{\columnwidth}{!}{
  \begin{tabular}{lccccc}
    \Xhline{1pt}
        
    \rowcolor[gray]{0.9} 
     & \multicolumn{3}{c}{\textbf{IFS}} & \multicolumn{2}{c}{\textbf{Overall}} \\
    \noalign{\vskip -2pt}\cmidrule(r){2-4}  \cmidrule(r){5-6}  \rowcolor[gray]{0.9} \noalign{\vskip -3pt}
    \multirow{-2}{*}{\textbf{Model}}   & \textbf{Expression} & \textbf{Variable} & \textbf{Length} & \textbf{IFS(AVG)} & \textbf{ARS}\\
    
    \hline\multicolumn{6}{c}{\textit{API Models}} \\ \hdashline
        GPT-4o & 0.087  $_{(3)}$ & \textbf{0.063}  $_{(1)}$ & \textbf{0.086}  $_{(1)}$ & \textbf{0.079}  $_{(1)}$ & \textbf{0.758}  $_{(1)}$  \\
        GPT-4 &\underline{0.066}  $_{(2)}$ & 0.101  $_{(8)}$ & 0.155  $_{(3)}$ & 0.107  $_{(3)}$ & \underline{0.738}  $_{(2)}$ \\

    \hline \multicolumn{6}{c}{\textit{Models Larger Than 20B Parameters}}\\ \hdashline
        Qwen2.5-72B-Inst.\textsuperscript{$\dagger$} &\textbf{0.065}  $_{(1)}$ & 0.076  $_{(3)}$ & 0.165  $_{(4)}$ & \underline{0.102}  $_{(2)}$ & 0.707  $_{(3)}$  \\
        C4AI-cmd-r-v01\textsuperscript{$\dagger$} &0.135  $_{(9)}$ & 0.082  $_{(5)}$ & \underline{0.143}  $_{(2)}$ & 0.120  $_{(4)}$ & 0.596  $_{(7)}$  \\
        Qwen2.5-32B-Inst.\textsuperscript{$\dagger$} &0.103  $_{(5)}$ & 0.087  $_{(6)}$ & 0.182  $_{(6)}$ & 0.124  $_{(5)}$ & 0.651  $_{(5)}$  \\
        Llama-3.1-70B-Inst. &0.101  $_{(4)}$ & \underline{0.076}  $_{(2)}$ & 0.263  $_{(13)}$ & 0.147  $_{(8)}$ & 0.694  $_{(4)}$  \\
        C4AI-cmd-r-08-2024\textsuperscript{$\dagger$} &0.114  $_{(7)}$ & 0.077  $_{(4)}$ & 0.238  $_{(11)}$ & 0.143  $_{(7)}$ & 0.626  $_{(6)}$  \\
        Qwen2.5-72B\textsuperscript{$\dagger$} &0.145  $_{(11)}$ & 0.094  $_{(7)}$ & 0.232  $_{(10)}$ & 0.157  $_{(11)}$ & 0.552  $_{(8)}$  \\
        Qwen2.5-32B\textsuperscript{$\dagger$} &0.196  $_{(15)}$ & 0.117  $_{(10)}$ & 0.685  $_{(19)}$ & 0.332  $_{(19)}$ & 0.396  $_{(16)}$ \\ 
        Llama-3.1-70B &0.225  $_{(18)}$ & 0.165  $_{(19)}$ & 0.380  $_{(17)}$ & 0.257  $_{(17)}$ & 0.433  $_{(14)}$  \\
    \hline\multicolumn{6}{c}{\textit{Models With 7-20B Parameters}}\\ \hdashline
        InternLM2.5-7b-chat-1m &0.107  $_{(6)}$ & 0.128  $_{(16)}$ & 0.193  $_{(7)}$ & 0.143  $_{(6)}$ & 0.533  $_{(10)}$  \\
        Qwen2.5-7B-Inst.\textsuperscript{$\dagger$} &0.142  $_{(10)}$ & 0.112  $_{(9)}$ & 0.250  $_{(12)}$ & 0.168  $_{(13)}$ & 0.520  $_{(11)}$  \\
        Qwen2.5-14B-Inst.\textsuperscript{$\dagger$} &0.133  $_{(8)}$ & 0.124  $_{(15)}$ & 0.206  $_{(9)}$ & 0.154  $_{(9)}$ & 0.548  $_{(9)}$  \\
        Llama-3.1-8B-Inst. &0.154  $_{(13)}$ & 0.120  $_{(12)}$ & 0.199  $_{(8)}$ & 0.158  $_{(12)}$ & 0.491  $_{(12)}$  \\
        GLM-4-9b-chat-1m &0.151  $_{(12)}$ & 0.138  $_{(18)}$ & 0.175  $_{(5)}$ & 0.155  $_{(10)}$ & 0.490  $_{(13)}$  \\
        Llama-3.1-8B &0.215  $_{(17)}$ & 0.119  $_{(11)}$ & 0.319  $_{(16)}$ & 0.217  $_{(15)}$ & 0.392  $_{(17)}$  \\
        Qwen2.5-14B\textsuperscript{$\dagger$} &0.171  $_{(14)}$ & 0.121  $_{(13)}$ & 0.515  $_{(18)}$ & 0.269  $_{(18)}$ & 0.386  $_{(18)}$  \\
        LWM-Text-Chat-1M &0.204  $_{(16)}$ & 0.130  $_{(17)}$ & 0.282  $_{(15)}$ & 0.206  $_{(14)}$ & 0.411  $_{(15)}$  \\
        LWM-Text-1M &0.237  $_{(19)}$ & 0.179  $_{(20)}$ & 0.280  $_{(14)}$ & 0.232  $_{(16)}$ & 0.205  $_{(20)}$  \\
        Qwen2.5-7B\textsuperscript{$\dagger$} &0.265  $_{(20)}$ & 0.123  $_{(14)}$ & 0.785  $_{(20)}$ & 0.391  $_{(20)}$ & 0.213  $_{(19)}$  \\ 
        \Xhline{1pt}
  \end{tabular}
  }
  \vspace{-1ex}
  \caption{Instruction Following Stability (IFS) from three perspectives, with rankings in parentheses (smaller is better). The overall order is based on average IFS rankings. $\dagger$ denotes truncation of context $X$ on the longest interval.}
  \label{tab:ifs}
  \vspace{-3.5ex}
\end{table}

\subsection{Results on \ours}
\subsectiontitle{Task-categorized  Performance}  
As shown in Table~\ref{tab:main results}, two closed-source models achieve the highest scores, with the top score reaching only 0.758.
This highlights substantial room for improvement in instruction-following capabilities.
Generally, larger parameter sizes are associated with better performance.
However, models with supervised instruction tuning often outperform their base counterparts, even when smaller in size.
For instance, InternLM2.5-7b-chat-1m scores 0.101 higher than Llama-3.1-70B, demonstrating that fine-tuning on instruction or conversational data significantly enhances performance. 
While closed-source models dominate most tasks, open-source models excel in only a few (e.g., \textit{Blur-ID} and \textit{Batch-label}). 

Moreover, compared to LSI, most of models perform worse on LOI and LBI tasks, likely due to LSI's closer alignment with their training data, highlighting the need to enhance their ability to handle complex instructions. 
Similarly, the model performance on LBE~(LOE) tasks is better than that on LBI~(LOI), indicating that referencing specific elements is easier than using non-semantic IDs.

\begin{figure}[t]
    \centering
    \includegraphics[width=0.9\linewidth]{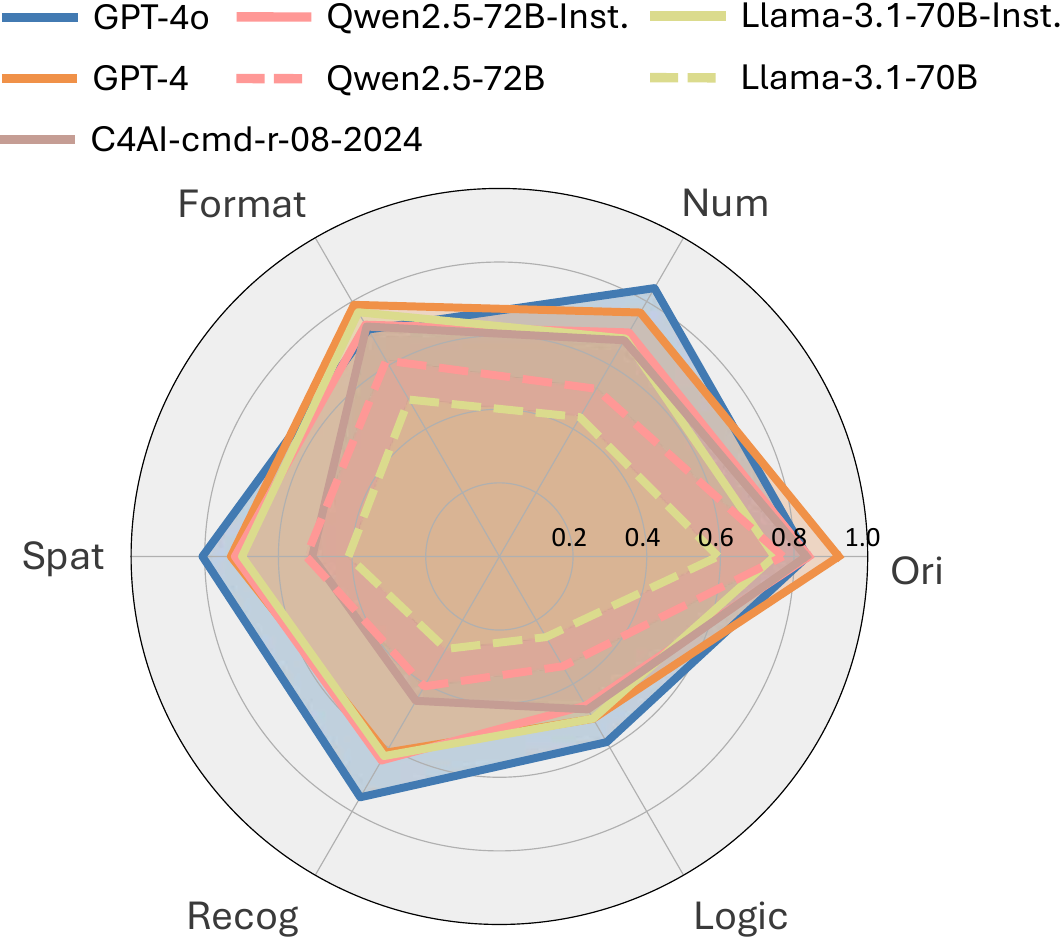}
    \vspace{-1ex}
    \caption{The Instruction Following Performance in six capabilities. Lines of the same color in the chart represent models from the same  series. Dashed lines represent base models, while solid lines represent their officially fine-tuned variants.}
    \label{fig:core-capabilities}
    \vspace{-3ex}
\end{figure}

\subsectiontitle{Capability-categorized Performance}
As shown in Figure \ref{fig:core-capabilities}, closed-source models lead in all dimensions. Within each model series, the solid lines consistently enclose their corresponding dashed lines, highlighting the significant impact of instruction fine-tuning on overall performance. Notably, the \textit{Format} dimension shows the tightest clustering, suggesting that formatting requirements are well-covered across training corpora for most models. 
In contrast, the \textit{Recog} dimension shows marked differences, likely reflecting varying levels of effort in incorporating numerical cognition data during training.

\begin{figure*}[t]
    \centering
    \includegraphics[width=0.95\linewidth]{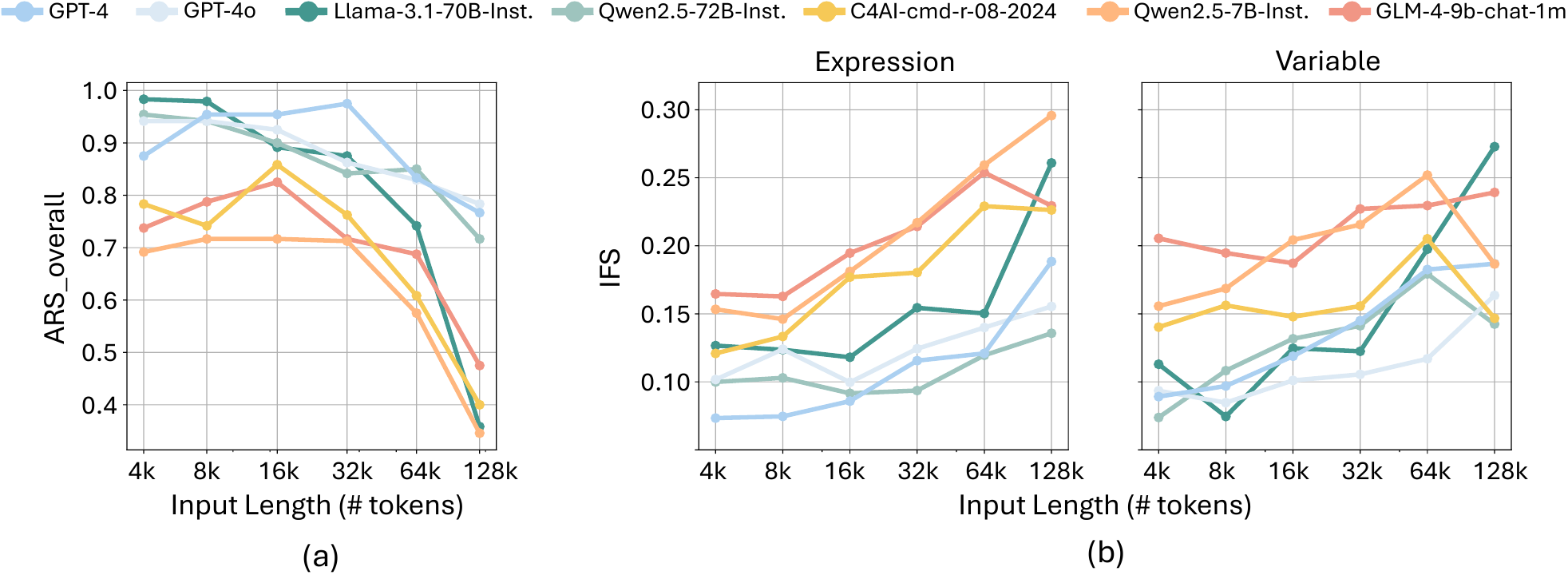}
    \vspace{-2ex}
    \caption{Overall ARS score (a) and instruction following stability (b) under different input length.}
    \label{fig:len1}
    \vspace{-3ex}
\end{figure*}

\subsectiontitle{Stability}
Table \ref{tab:ifs} presents the IFS scores and rankings from three perspectives, revealing discrepancies between model stability and task completion ability. 
For instance, while Qwen2.5-72B-Inst. underwent truncation, potentially affecting its stability in "Length", it still outperformed GPT-4, which had a higher ARS score. 
Furthermore, models exhibited distinct strengths and weaknesses across perspectives: GPT-4o showed less stability in "Expression", while GPT-4 struggled with instruction variables where Llama-3.1-70B-Inst. excelled. 
Instruction fine-tuning generally improved stability , but larger parameter size did not guarantee better performance, as Qwen2.5-72B-Inst. surpassed all closed-source models in "Expression".

\begin{figure}[!t]
    \centering
    \includegraphics[width=0.98\linewidth]{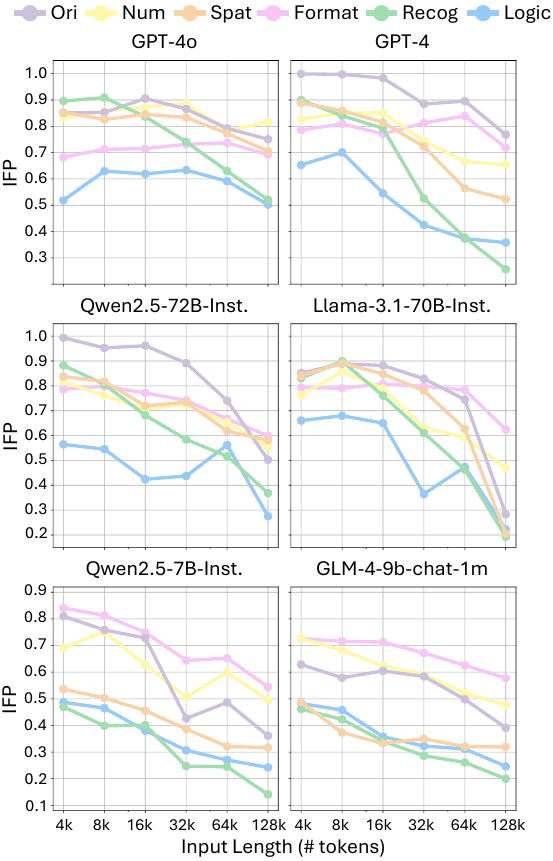}
    \vspace{-2ex}
    \caption{Instruction following performance in six core capabilities under different input length.}
    \label{fig:len2}
    \vspace{-3ex}
\end{figure}

\begin{figure*}[t]
    \centering
    \includegraphics[width=0.90\linewidth]{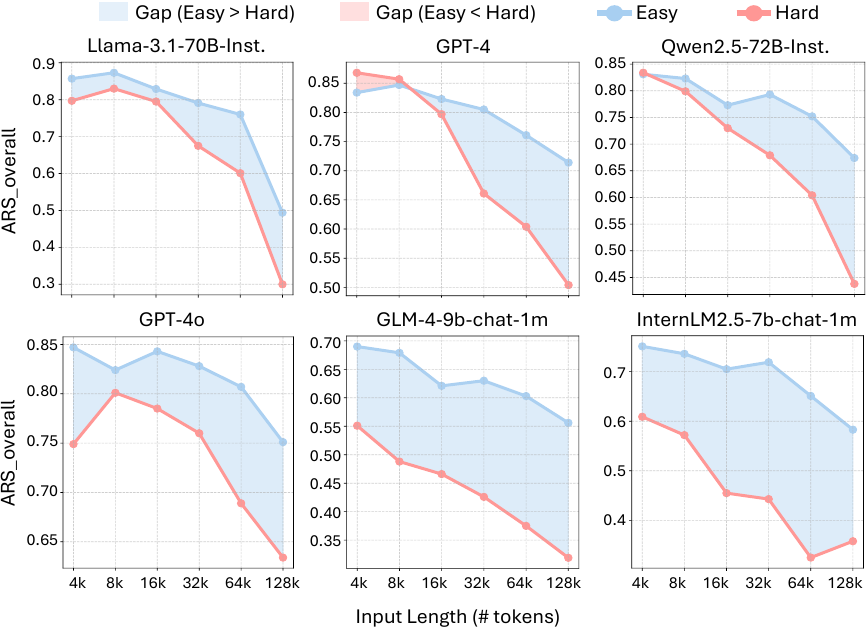}
    \vspace{-1ex}
    \caption{The impact of increasing context length on ARS scores in tasks with different instruction complexities.}
    \label{fig:performance-comparison}
    \vspace{-2ex}
\end{figure*}

\subsection{Effects of longer inputs}
\subsectiontitle{Overall Performance and Stability}
As shown in Figure \ref{fig:len1}(a), the performance of most models declines significantly as input length increases, particularly beyond 16k or 32k tokens. However, the rates of decline vary across models.
For instance, GPT-4o maintains relatively high scores in long-context scenarios, whereas Llama-3-70B-Inst. experiences a sharp drop, indicating its limited ability to handle extended inputs. 

The negative impact of input length is evident in stability metrics. As illustrated in Figure \ref{fig:len1}(b), most models demonstrate poorer stability in long-context scenarios, with the lowest stability observed at the longest input lengths. Interestingly, some models, such as C4AI-cmd-r-08-2024 and GLM-4-9b-chat-1m, exhibit the least stability at mid-range input lengths (e.g., 64k tokens), diverging from their overall performance trends. Additionally, the sensitivity to input length also varies across different perspectives. For instance, models like C4AI-cmd-r-08-2024 and Qwen2.5-7B-Inst. show more significant upward trend in "Expression" compared to "Variable". This highlights potential areas for improvement in enhancing instruction-following stability under long input contexts.

\subsectiontitle{Performance in six core capabilities}
Figure \ref{fig:len2} demonstrates the declining trends of six core capabilities across varying model sizes as input length increases. Notably, "Format" performance remains relatively stable across all input lengths in most models, suggesting that tasks related to formatting are less sensitive to longer contexts. Conversely, "Recog" experiences the steepest decline, highlighting the challenges models face in maintaining recognition ability as input length grows.

Model size also plays a crucial role in long-context instruction-following. Larger models generally are better at capabilities like "Ori," "Num," and "Spat," indicating that handling original content and spatial reasoning demands more model parameters. In contrast, format-related abilities are effectively managed even by smaller models, suggesting they are less dependent on model size.

\subsection{Impact of Instruction Complexity}
To analyze how context length and instruction complexity simultaneously affect model performance, we divide tasks into two groups based on scoring weight: easy ($\leq$10) and hard (\textgreater  10). 
As noted in Section \ref{sec:ars}, higher weights indicate greater challenges, with 10 chosen to balance the number of tasks in each group. 


As shown in Figure \ref{fig:performance-comparison}, the performance of different models varies significantly. Llama-3.1-70B-Inst. struggles with long contexts across all tasks, highlighting its difficulty in handling long inputs. For GPT-4 and Qwen2.5-72B-Inst., the performance on both groups is similar in short contexts, but has sharper declines on hard tasks with longer contexts, revealing their limitations in handling complex tasks in long-context scenarios. Other models exhibit larger performance gaps even in short input (i.e., 4k tokens), indicating greater task difficulty impact for them. Generally, hard group degrades more significantly as context length increases, suggesting a compounded negative effect of task complexity and context length on LLM performance.

\section{Conclusion}
In this work, we systematically evaluate LLMs' instruction-following capabilities and stability in long-context scenarios. We develop \ours, a benchmark that encompasses three long-context scenarios and a diverse set of 11 tasks, complemented by a method for instruction expansion across three distinct perspectives. For evaluation, we introduce \ourseval, an automated rubric-based scoring method, enabling fast and accurate assessments of model task performance and stability. Based on the benchmark and scoring method, we conduct extensive experiments on 20 prominent LLMs, revealing significant room for improvement in instruction-following capabilities and stability, especially in long-context scenarios.

\section*{Limitation}
Our study has several limitations. 
Firstly, due to constraints in programmatic validation, our task scenarios lack comprehensive support for semantic constraints, necessitating future improvements for better validation. 
Secondly, the inference process for very long inputs requires significant computational resources and time, limiting the scale of our dataset and potentially affecting the reproducibility of our work. As a result, our benchmark example comprises fewer than 3,000 samples, tested across only three perspectives.
In fact, many unexplored aspects of stability remain, such as LLMs' consistency in handling input formatting and context domain shifts. Additionally, larger datasets can enable more statistically significant conclusions. We encourage the community to use our proposed protocol to expand the evaluation set and conduct more extensive analyzes.

Finally, while \ourseval enables efficient and automated evaluations, the reliability of its results depends heavily on the design of the scoring rubric and the implementation of the evaluation programs. These components require significant time and effort to prepare before conducting evaluations. We leave the automation or reduction of these manual efforts for future work.

\section*{Acknowledgments}
This work was supported in part by National Natural Science Foundation of China (No. 92270119) and Key Laboratory of Advanced Theory and Application in Statistics and Data Science, Ministry of Education.


\bibliography{main}

\begin{thebibliography}{57}
\providecommand{\natexlab}[1]{#1}

\bibitem[{Achiam et~al.(2023)Achiam, Adler, Agarwal, Ahmad, Akkaya, Aleman, Almeida, Altenschmidt, Altman, Anadkat et~al.}]{achiam2023gpt}
Josh Achiam, Steven Adler, Sandhini Agarwal, Lama Ahmad, Ilge Akkaya, Florencia~Leoni Aleman, Diogo Almeida, Janko Altenschmidt, Sam Altman, Shyamal Anadkat, et~al. 2023.
\newblock Gpt-4 technical report.
\newblock \emph{arXiv preprint arXiv:2303.08774}.

\bibitem[{An et~al.(2023)An, Gong, Zhong, Zhao, Li, Zhang, Kong, and Qiu}]{an2023leval}
Chenxin An, Shansan Gong, Ming Zhong, Xingjian Zhao, Mukai Li, Jun Zhang, Lingpeng Kong, and Xipeng Qiu. 2023.
\newblock L-eval: Instituting standardized evaluation for long context language models.
\newblock \emph{arXiv preprint arXiv:2307.11088}.

\bibitem[{Anthropic(2024)}]{anthropic2024claude}
AI~Anthropic. 2024.
\newblock The claude 3 model family: Opus, sonnet, haiku.
\newblock \emph{Claude-3 Model Card}, 1.

\bibitem[{Bai et~al.(2023)Bai, Lv, Zhang, Lyu, Tang, Huang, Du, Liu, Zeng, Hou et~al.}]{bai2023longbench}
Yushi Bai, Xin Lv, Jiajie Zhang, Hongchang Lyu, Jiankai Tang, Zhidian Huang, Zhengxiao Du, Xiao Liu, Aohan Zeng, Lei Hou, et~al. 2023.
\newblock Longbench: A bilingual, multitask benchmark for long context understanding.
\newblock \emph{arXiv preprint arXiv:2308.14508}.

\bibitem[{Brown(1998)}]{brown1998cv}
Charles~E Brown. 1998.
\newblock Coefficient of variation.
\newblock In \emph{Applied multivariate statistics in geohydrology and related sciences}, pages 155--157. Springer.

\bibitem[{Brown(2020)}]{brown2020language}
Tom~B Brown. 2020.
\newblock Language models are few-shot learners.
\newblock \emph{arXiv preprint arXiv:2005.14165}.

\bibitem[{Cai et~al.(2024)Cai, Cao, Chen, Chen, Chen, Chen, Chen, Chen, Chen, Chu et~al.}]{cai2024InternLM2}
Zheng Cai, Maosong Cao, Haojiong Chen, Kai Chen, Keyu Chen, Xin Chen, Xun Chen, Zehui Chen, Zhi Chen, Pei Chu, et~al. 2024.
\newblock Internlm2 technical report.
\newblock \emph{arXiv preprint arXiv:2403.17297}.

\bibitem[{Chen et~al.(2024{\natexlab{a}})Chen, Xiao, Zhang, Luo, Lian, and Liu}]{chen2024bge}
Jianlv Chen, Shitao Xiao, Peitian Zhang, Kun Luo, Defu Lian, and Zheng Liu. 2024{\natexlab{a}}.
\newblock Bge m3-embedding: Multi-lingual, multi-functionality, multi-granularity text embeddings through self-knowledge distillation.
\newblock \emph{arXiv preprint arXiv:2402.03216}.

\bibitem[{Chen et~al.(2024{\natexlab{b}})Chen, Chen, Zhou, He, and He}]{DBLP:conf/emnlp/ChenCZH024}
Kedi Chen, Qin Chen, Jie Zhou, Yishen He, and Liang He. 2024{\natexlab{b}}.
\newblock \href {https://aclanthology.org/2024.findings-emnlp.529} {Diahalu: {A} dialogue-level hallucination evaluation benchmark for large language models}.
\newblock In \emph{Findings of the Association for Computational Linguistics: {EMNLP} 2024, Miami, Florida, USA, November 12-16, 2024}, pages 9057--9079. Association for Computational Linguistics.

\bibitem[{Chen et~al.(2025)Chen, Lei, Zhang, Zhang, Chen, Zhou, He, Guo, Chen, and Zhang}]{DBLP:journals/corr/abs-2503-13109}
Kedi Chen, Zhikai Lei, Fan Zhang, Yinqi Zhang, Qin Chen, Jie Zhou, Liang He, Qipeng Guo, Kai Chen, and Wei Zhang. 2025.
\newblock \href {https://doi.org/10.48550/ARXIV.2503.13109} {Code-driven inductive synthesis: Enhancing reasoning abilities of large language models with sequences}.
\newblock \emph{CoRR}, abs/2503.13109.

\bibitem[{Chowdhery et~al.(2023)Chowdhery, Narang, Devlin, Bosma, Mishra, Roberts, Barham, Chung, Sutton, Gehrmann et~al.}]{chowdhery2023palm}
Aakanksha Chowdhery, Sharan Narang, Jacob Devlin, Maarten Bosma, Gaurav Mishra, Adam Roberts, Paul Barham, Hyung~Won Chung, Charles Sutton, Sebastian Gehrmann, et~al. 2023.
\newblock Palm: Scaling language modeling with pathways.
\newblock \emph{Journal of Machine Learning Research}, 24(240):1--113.

\bibitem[{{Cohere For AI}(2024)}]{cohere_for_ai_2024}
{Cohere For AI}. 2024.
\newblock \href {https://doi.org/10.57967/hf/3134} {c4ai-command-r-08-2024}.

\bibitem[{Cook et~al.(2024)Cook, Rockt{\"a}schel, Foerster, Aumiller, and Wang}]{cook2024ticking}
Jonathan Cook, Tim Rockt{\"a}schel, Jakob Foerster, Dennis Aumiller, and Alex Wang. 2024.
\newblock Ticking all the boxes: Generated checklists improve llm evaluation and generation.
\newblock \emph{arXiv preprint arXiv:2410.03608}.

\bibitem[{Dong et~al.(2024)Dong, Tang, Li, Zhao, and Wen}]{dong2024bamboo}
Zican Dong, Tianyi Tang, Junyi Li, Wayne~Xin Zhao, and Ji-Rong Wen. 2024.
\newblock Bamboo: A comprehensive benchmark for evaluating long text modeling capacities of large language models.
\newblock In \emph{Proceedings of the 2024 Joint International Conference on Computational Linguistics, Language Resources and Evaluation (LREC-COLING 2024)}, pages 2086--2099.

\bibitem[{Dubey et~al.(2024)Dubey, Jauhri, Pandey, Kadian, Al-Dahle, Letman, Mathur, Schelten, Yang, Fan et~al.}]{dubey2024llama}
Abhimanyu Dubey, Abhinav Jauhri, Abhinav Pandey, Abhishek Kadian, Ahmad Al-Dahle, Aiesha Letman, Akhil Mathur, Alan Schelten, Amy Yang, Angela Fan, et~al. 2024.
\newblock The llama 3 herd of models.
\newblock \emph{arXiv preprint arXiv:2407.21783}.

\bibitem[{Gavin et~al.(2024)Gavin, Zheng, Liu, Que, Wang, Yang, Zhang, Huang, Chen, and Zhang}]{Gavin2024LongInsAC}
Shawn Gavin, Tuney Zheng, Jiaheng Liu, Quehry Que, Noah Wang, Jian Yang, Chenchen Zhang, Wenhao Huang, Wenhu Chen, and Ge~Zhang. 2024.
\newblock \href {https://api.semanticscholar.org/CorpusID:270710851} {Longins: A challenging long-context instruction-based exam for llms}.
\newblock \emph{ArXiv}, abs/2406.17588.

\bibitem[{GLM et~al.(2024)GLM, Zeng, Xu, Wang, Zhang, Yin, Rojas, Feng, Zhao, Lai et~al.}]{glm2024chatglm}
Team GLM, Aohan Zeng, Bin Xu, Bowen Wang, Chenhui Zhang, Da~Yin, Diego Rojas, Guanyu Feng, Hanlin Zhao, Hanyu Lai, et~al. 2024.
\newblock Chatglm: A family of large language models from glm-130b to glm-4 all tools.
\newblock \emph{arXiv preprint arXiv:2406.12793}.

\bibitem[{{Greg Kamradt}(2023)}]{gkamradt2023llmtest}
{Greg Kamradt}. 2023.
\newblock {LLMTest: Needle in a Haystack}.
\newblock {\url{https://github.com/gkamradt/LLMTest_NeedleInAHaystack}}.

\bibitem[{He et~al.(2024)He, Zeng, Huang, Chen, Xiao, He, Zhou, Liang, and Xiao}]{he2024can}
Qianyu He, Jie Zeng, Wenhao Huang, Lina Chen, Jin Xiao, Qianxi He, Xunzhe Zhou, Jiaqing Liang, and Yanghua Xiao. 2024.
\newblock Can large language models understand real-world complex instructions?
\newblock In \emph{Proceedings of the AAAI Conference on Artificial Intelligence}, volume~38, pages 18188--18196.

\bibitem[{Honovich et~al.(2023)Honovich, Scialom, Levy, and Schick}]{honovich2023unnatural}
Or~Honovich, Thomas Scialom, Omer Levy, and Timo Schick. 2023.
\newblock Unnatural instructions: Tuning language models with (almost) no human labor.
\newblock In \emph{Proceedings of the 61st Annual Meeting of the Association for Computational Linguistics (Volume 1: Long Papers)}, pages 14409--14428.

\bibitem[{Hsieh et~al.(2024)Hsieh, Sun, Kriman, Acharya, Rekesh, Jia, and Ginsburg}]{hsieh2024ruler}
Cheng-Ping Hsieh, Simeng Sun, Samuel Kriman, Shantanu Acharya, Dima Rekesh, Fei Jia, and Boris Ginsburg. 2024.
\newblock Ruler: What's the real context size of your long-context language models?
\newblock \emph{arXiv preprint arXiv:2404.06654}.

\bibitem[{Huang et~al.(2021)Huang, Cao, Parulian, Ji, and Wang}]{huang2021efficientgovreport}
Luyang Huang, Shuyang Cao, Nikolaus Parulian, Heng Ji, and Lu~Wang. 2021.
\newblock Efficient attentions for long document summarization.
\newblock \emph{arXiv preprint arXiv:2104.02112}.

\bibitem[{Jiang et~al.(2023)Jiang, Wang, Zeng, Zhong, Li, Mi, Shang, Jiang, Liu, and Wang}]{jiang2023followbench}
Yuxin Jiang, Yufei Wang, Xingshan Zeng, Wanjun Zhong, Liangyou Li, Fei Mi, Lifeng Shang, Xin Jiang, Qun Liu, and Wei Wang. 2023.
\newblock Followbench: A multi-level fine-grained constraints following benchmark for large language models.
\newblock \emph{arXiv preprint arXiv:2310.20410}.

\bibitem[{Kwan et~al.(2023)Kwan, Zeng, Wang, Sun, Li, Shang, Liu, and Wong}]{kwan2023m4le}
Wai-Chung Kwan, Xingshan Zeng, Yufei Wang, Yusen Sun, Liangyou Li, Lifeng Shang, Qun Liu, and Kam-Fai Wong. 2023.
\newblock M4le: A multi-ability multi-range multi-task multi-domain long-context evaluation benchmark for large language models.
\newblock \emph{arXiv preprint arXiv:2310.19240}.

\bibitem[{Kwon et~al.(2023)Kwon, Li, Zhuang, Sheng, Zheng, Yu, Gonzalez, Zhang, and Stoica}]{kwon2023efficient}
Woosuk Kwon, Zhuohan Li, Siyuan Zhuang, Ying Sheng, Lianmin Zheng, Cody~Hao Yu, Joseph Gonzalez, Hao Zhang, and Ion Stoica. 2023.
\newblock Efficient memory management for large language model serving with pagedattention.
\newblock In \emph{Proceedings of the 29th Symposium on Operating Systems Principles}, pages 611--626.

\bibitem[{Levy et~al.(2024)Levy, Jacoby, and Goldberg}]{levy2024same}
Mosh Levy, Alon Jacoby, and Yoav Goldberg. 2024.
\newblock Same task, more tokens: the impact of input length on the reasoning performance of large language models.
\newblock \emph{arXiv preprint arXiv:2402.14848}.

\bibitem[{Li et~al.(2023)Li, Wang, Zheng, and Zhang}]{li2023loogle}
Jiaqi Li, Mengmeng Wang, Zilong Zheng, and Muhan Zhang. 2023.
\newblock Loogle: Can long-context language models understand long contexts?
\newblock \emph{arXiv preprint arXiv:2311.04939}.

\bibitem[{Li et~al.(2024{\natexlab{a}})Li, Zhang, Liu, and Chen}]{li2024needlebench}
Mo~Li, Songyang Zhang, Yunxin Liu, and Kai Chen. 2024{\natexlab{a}}.
\newblock Needlebench: Can llms do retrieval and reasoning in 1 million context window?
\newblock \emph{arXiv preprint arXiv:2407.11963}.

\bibitem[{Li et~al.(2024{\natexlab{b}})Li, Cui, Zhao, Kong, and Bi}]{li2024gsm}
Qintong Li, Leyang Cui, Xueliang Zhao, Lingpeng Kong, and Wei Bi. 2024{\natexlab{b}}.
\newblock Gsm-plus: A comprehensive benchmark for evaluating the robustness of llms as mathematical problem solvers.
\newblock \emph{arXiv preprint arXiv:2402.19255}.

\bibitem[{Li et~al.(2024{\natexlab{c}})Li, Zhang, Do, Yue, and Chen}]{li2024long}
Tianle Li, Ge~Zhang, Quy~Duc Do, Xiang Yue, and Wenhu Chen. 2024{\natexlab{c}}.
\newblock Long-context llms struggle with long in-context learning.
\newblock \emph{arXiv preprint arXiv:2404.02060}.

\bibitem[{Liu et~al.(2024{\natexlab{a}})Liu, Yan, Zaharia, and Abbeel}]{Liu2024WorldMO}
Hao Liu, Wilson Yan, Matei Zaharia, and Pieter Abbeel. 2024{\natexlab{a}}.
\newblock \href {https://api.semanticscholar.org/CorpusID:267637090} {World model on million-length video and language with blockwise ringattention}.
\newblock \emph{ArXiv}, abs/2402.08268.

\bibitem[{Liu et~al.(2024{\natexlab{b}})Liu, Lin, Hewitt, Paranjape, Bevilacqua, Petroni, and Liang}]{liu2024lost}
Nelson~F Liu, Kevin Lin, John Hewitt, Ashwin Paranjape, Michele Bevilacqua, Fabio Petroni, and Percy Liang. 2024{\natexlab{b}}.
\newblock Lost in the middle: How language models use long contexts.
\newblock \emph{Transactions of the Association for Computational Linguistics}, 12:157--173.

\bibitem[{Lu et~al.(2024)Lu, Bansal, Xia, Liu, Li, Hajishirzi, Cheng, Chang, Galley, and Gao}]{lu2023mathvista}
Pan Lu, Hritik Bansal, Tony Xia, Jiacheng Liu, Chunyuan Li, Hannaneh Hajishirzi, Hao Cheng, Kai-Wei Chang, Michel Galley, and Jianfeng Gao. 2024.
\newblock Mathvista: Evaluating mathematical reasoning of foundation models in visual contexts.
\newblock In \emph{International Conference on Learning Representations (ICLR)}.

\bibitem[{Macqueen(1967)}]{macqueen1967kmeans}
J~Macqueen. 1967.
\newblock Some methods for classification and analysis of multivariate observations.
\newblock In \emph{Proceedings of 5-th Berkeley Symposium on Mathematical Statistics and Probability/University of California Press}.

\bibitem[{Narayan et~al.(2018)Narayan, Cohen, and Lapata}]{narayan2018donxsum}
Shashi Narayan, Shay~B Cohen, and Mirella Lapata. 2018.
\newblock Don't give me the details, just the summary! topic-aware convolutional neural networks for extreme summarization.
\newblock \emph{arXiv preprint arXiv:1808.08745}.

\bibitem[{Ni et~al.(2024)Ni, Cai, Wei, Wang, Yin, and Li}]{ni2024xl}
Xuanfan Ni, Hengyi Cai, Xiaochi Wei, Shuaiqiang Wang, Dawei Yin, and Piji Li. 2024.
\newblock Xl\textsuperscript{2} bench: A benchmark for extremely long context understanding with long-range dependencies.
\newblock \emph{arXiv preprint arXiv:2404.05446}.

\bibitem[{O’Connor and Andreas(2021)}]{o2021context}
Joe O’Connor and Jacob Andreas. 2021.
\newblock What context features can transformer language models use?
\newblock In \emph{Proceedings of the 59th Annual Meeting of the Association for Computational Linguistics and the 11th International Joint Conference on Natural Language Processing (Volume 1: Long Papers)}, pages 851--864.

\bibitem[{Parmar et~al.(2024)Parmar, Patel, Varshney, Nakamura, Luo, Mashetty, Mitra, and Baral}]{parmar2024logicbench}
Mihir Parmar, Nisarg Patel, Neeraj Varshney, Mutsumi Nakamura, Man Luo, Santosh Mashetty, Arindam Mitra, and Chitta Baral. 2024.
\newblock Logicbench: Towards systematic evaluation of logical reasoning ability of large language models.
\newblock In \emph{Proceedings of the 62nd Annual Meeting of the Association for Computational Linguistics (Volume 1: Long Papers)}, pages 13679--13707.

\bibitem[{Peng et~al.(2023)Peng, Quesnelle, Fan, and Shippole}]{Peng2023YaRNEC}
Bowen Peng, Jeffrey Quesnelle, Honglu Fan, and Enrico Shippole. 2023.
\newblock \href {https://api.semanticscholar.org/CorpusID:261493986} {Yarn: Efficient context window extension of large language models}.
\newblock \emph{ArXiv}, abs/2309.00071.

\bibitem[{Qin et~al.(2024)Qin, Song, Hu, Yao, Cho, Wang, Wu, Liu, Liu, and Yu}]{qin2024infobench}
Yiwei Qin, Kaiqiang Song, Yebowen Hu, Wenlin Yao, Sangwoo Cho, Xiaoyang Wang, Xuansheng Wu, Fei Liu, Pengfei Liu, and Dong Yu. 2024.
\newblock Infobench: Evaluating instruction following ability in large language models.
\newblock \emph{arXiv preprint arXiv:2401.03601}.

\bibitem[{Que et~al.(2024)Que, Duan, He, Mou, Zhou, Liu, Rong, Wang, Yang, Zhang et~al.}]{que2024hellobench}
Haoran Que, Feiyu Duan, Liqun He, Yutao Mou, Wangchunshu Zhou, Jiaheng Liu, Wenge Rong, Zekun~Moore Wang, Jian Yang, Ge~Zhang, et~al. 2024.
\newblock Hellobench: Evaluating long text generation capabilities of large language models.
\newblock \emph{arXiv preprint arXiv:2409.16191}.

\bibitem[{Sakai et~al.(2024)Sakai, Nohejl, Hang, Kamigaito, and Watanabe}]{Sakai2024TowardTE}
Yusuke Sakai, Adam Nohejl, Jiangnan Hang, Hidetaka Kamigaito, and Taro Watanabe. 2024.
\newblock Toward the evaluation of large language models considering score variance across instruction templates.
\newblock In \emph{Proceedings of the 7th BlackboxNLP Workshop: Analyzing and Interpreting Neural Networks for NLP}, pages 499--529.

\bibitem[{Shaham et~al.(2023)Shaham, Ivgi, Efrat, Berant, and Levy}]{shaham2023zeroscrolls}
Uri Shaham, Maor Ivgi, Avia Efrat, Jonathan Berant, and Omer Levy. 2023.
\newblock Zeroscrolls: A zero-shot benchmark for long text understanding.
\newblock In \emph{Findings of the Association for Computational Linguistics: EMNLP 2023}, pages 7977--7989.

\bibitem[{Shaham et~al.(2022)Shaham, Segal, Ivgi, Efrat, Yoran, Haviv, Gupta, Xiong, Geva, Berant et~al.}]{shaham2022scrolls}
Uri Shaham, Elad Segal, Maor Ivgi, Avia Efrat, Ori Yoran, Adi Haviv, Ankit Gupta, Wenhan Xiong, Mor Geva, Jonathan Berant, et~al. 2022.
\newblock Scrolls: Standardized comparison over long language sequences.
\newblock In \emph{Proceedings of the 2022 Conference on Empirical Methods in Natural Language Processing}, pages 12007--12021.

\bibitem[{Shechtman(2013)}]{shechtman2013coefficient}
Orit Shechtman. 2013.
\newblock The coefficient of variation as an index of measurement reliability.
\newblock In \emph{Methods of clinical epidemiology}, pages 39--49. Springer.

\bibitem[{Tan et~al.(2024)Tan, Guo, Shi, Xu, Liu, Feng, Li, Wang, Shang, Liu et~al.}]{tan2024proxyqa}
Haochen Tan, Zhijiang Guo, Zhan Shi, Lu~Xu, Zhili Liu, Yunlong Feng, Xiaoguang Li, Yasheng Wang, Lifeng Shang, Qun Liu, et~al. 2024.
\newblock Proxyqa: An alternative framework for evaluating long-form text generation with large language models.
\newblock \emph{arXiv preprint arXiv:2401.15042}.

\bibitem[{Taori et~al.(2023)Taori, Gulrajani, Zhang, Dubois, Li, Guestrin, Liang, and Hashimoto}]{alpaca}
Rohan Taori, Ishaan Gulrajani, Tianyi Zhang, Yann Dubois, Xuechen Li, Carlos Guestrin, Percy Liang, and Tatsunori~B. Hashimoto. 2023.
\newblock Stanford alpaca: An instruction-following llama model.
\newblock \url{https://github.com/tatsu-lab/stanford_alpaca}.

\bibitem[{Wang et~al.(2024)Wang, Chen, Fu, Liao, Zhang, Wu, Yu, Xu, Zhang, Luo et~al.}]{wang2024leave}
Minzheng Wang, Longze Chen, Cheng Fu, Shengyi Liao, Xinghua Zhang, Bingli Wu, Haiyang Yu, Nan Xu, Lei Zhang, Run Luo, et~al. 2024.
\newblock Leave no document behind: Benchmarking long-context llms with extended multi-doc qa.
\newblock \emph{arXiv preprint arXiv:2406.17419}.

\bibitem[{Wang et~al.(2023)Wang, Li, Chen, Cai, Zhu, Lin, Cao, Liu, Liu, and Sui}]{wang2023largebias}
Peiyi Wang, Lei Li, Liang Chen, Zefan Cai, Dawei Zhu, Binghuai Lin, Yunbo Cao, Qi~Liu, Tianyu Liu, and Zhifang Sui. 2023.
\newblock Large language models are not fair evaluators.
\newblock \emph{arXiv preprint arXiv:2305.17926}.

\bibitem[{Wen et~al.(2024)Wen, Ke, Gu, Wu, Huang, Zhou, Li, Hu, Gao, Xu et~al.}]{wen2024complexbench}
Bosi Wen, Pei Ke, Xiaotao Gu, Lindong Wu, Hao Huang, Jinfeng Zhou, Wenchuang Li, Binxin Hu, Wendy Gao, Jiaxin Xu, et~al. 2024.
\newblock Benchmarking complex instruction-following with multiple constraints composition.
\newblock \emph{arXiv preprint arXiv:2407.03978}.

\bibitem[{Yang et~al.(2024)Yang, Yang, Hui, Zheng, Yu, Zhou, Li, Li, Liu, Huang et~al.}]{qwen2}
An~Yang, Baosong Yang, Binyuan Hui, Bo~Zheng, Bowen Yu, Chang Zhou, Chengpeng Li, Chengyuan Li, Dayiheng Liu, Fei Huang, et~al. 2024.
\newblock Qwen2 technical report.
\newblock \emph{arXiv preprint arXiv:2407.10671}.

\bibitem[{Zhang et~al.(2024{\natexlab{a}})Zhang, Shen, Luo, Zhang, Liang, Yang, Lin, Qiao, Chen, Cui et~al.}]{zhang2024cfbench}
Tao Zhang, Yanjun Shen, Wenjing Luo, Yan Zhang, Hao Liang, Fan Yang, Mingan Lin, Yujing Qiao, Weipeng Chen, Bin Cui, et~al. 2024{\natexlab{a}}.
\newblock Cfbench: A comprehensive constraints-following benchmark for llms.
\newblock \emph{arXiv preprint arXiv:2408.01122}.

\bibitem[{Zhang et~al.(2024{\natexlab{b}})Zhang, Ladhak, Durmus, Liang, Mckeown, and Hashimoto}]{zhang2024benchmarking}
Tianyi Zhang, Faisal Ladhak, Esin Durmus, Percy Liang, Kathleen Mckeown, and Tatsunori~B Hashimoto. 2024{\natexlab{b}}.
\newblock Benchmarking large language models for news summarization.
\newblock \emph{Transactions of the Association for Computational Linguistics}, 11:39--57.

\bibitem[{Zhang et~al.(2024{\natexlab{c}})Zhang, Chen, Hu, Xu, Chen, Hao, Han, Thai, Wang, Liu et~al.}]{zhang2024infinitebench}
Xinrong Zhang, Yingfa Chen, Shengding Hu, Zihang Xu, Junhao Chen, Moo Hao, Xu~Han, Zhen Thai, Shuo Wang, Zhiyuan Liu, et~al. 2024{\natexlab{c}}.
\newblock $\infty$ bench: Extending long context evaluation beyond 100k tokens.
\newblock In \emph{Proceedings of the 62nd Annual Meeting of the Association for Computational Linguistics (Volume 1: Long Papers)}, pages 15262--15277.

\bibitem[{Zhong et~al.(2021)Zhong, Yin, Yu, Zaidi, Mutuma, Jha, Awadallah, Celikyilmaz, Liu, Qiu et~al.}]{zhong2021qmsum}
Ming Zhong, Da~Yin, Tao Yu, Ahmad Zaidi, Mutethia Mutuma, Rahul Jha, Ahmed~Hassan Awadallah, Asli Celikyilmaz, Yang Liu, Xipeng Qiu, et~al. 2021.
\newblock Qmsum: A new benchmark for query-based multi-domain meeting summarization.
\newblock \emph{arXiv preprint arXiv:2104.05938}.

\bibitem[{Zhou et~al.(2023{\natexlab{a}})Zhou, Lu, Mishra, Brahma, Basu, Luan, Zhou, and Hou}]{ifezhou2023instruction}
Jeffrey Zhou, Tianjian Lu, Swaroop Mishra, Siddhartha Brahma, Sujoy Basu, Yi~Luan, Denny Zhou, and Le~Hou. 2023{\natexlab{a}}.
\newblock Instruction-following evaluation for large language models.
\newblock \emph{arXiv preprint arXiv:2311.07911}.

\bibitem[{Zhou et~al.(2023{\natexlab{b}})Zhou, Lu, Mishra, Brahma, Basu, Luan, Zhou, and Hou}]{zhou2023instruction}
Jeffrey Zhou, Tianjian Lu, Swaroop Mishra, Siddhartha Brahma, Sujoy Basu, Yi~Luan, Denny Zhou, and Le~Hou. 2023{\natexlab{b}}.
\newblock Instruction-following evaluation for large language models.
\newblock \emph{arXiv preprint arXiv:2311.07911}.

\end{thebibliography}
\clearpage
\newpage
\appendix
\section{Details in Data Collection}
\label{appdix:datacollection}
\begin{figure*}[t]
    \centering
    \includegraphics[width=0.95\linewidth]{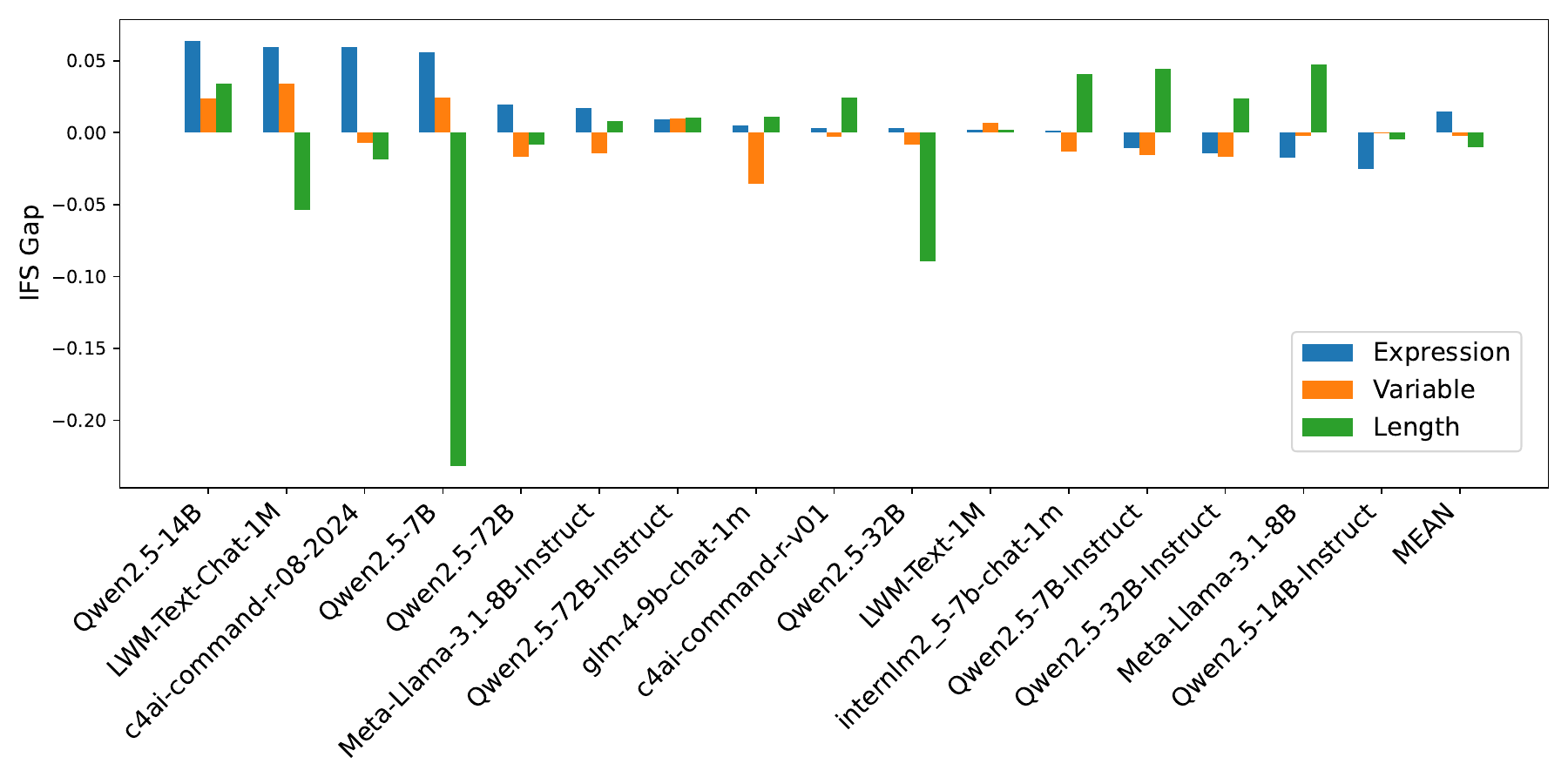}
    \vspace{-2.5ex}
    \caption{IFS Gap Across Three Perspectives with RECS vs. Random Sampling. Positive values indicate an IFS increase with RECS compared to Random Sampling, while negative values indicate a decrease. "MEAN" on the x-axis represents the average IFS gap.}
    \label{fig:gap}
    \vspace{-2ex}
\end{figure*}

\subsection{List}
\label{appdix:list detail}
To ensure the quality of tasks, the context $X$ for the List scenario is constructed as a ordered list, consisting of two types of elements: randomly generated UUIDs and natural language instruction texts. \cite{liu2024lost} used UUIDs to build key-value pairs and explained how large models utilize input context. Inspired by this work, we generated a set of unique 128-bit UUIDs as the first part of the list. However, real-world scenarios often involve some level of semantic noise, and transformer-based language models may exhibit varying sensitivities to different linguistic features in their input~\cite{o2021context}. Therefore, to enhance complexity and realism, we selected a subset of instruction texts from the Alpaca-52k dataset \cite{alpaca}, which serve as the second type of elements in the list. 
We found that when instruction texts are mixed into the retrieval list, the model's attention tends to be drawn to the embedded instructions, leading to a prioritization of following them rather than focusing on the originally assigned task. As a result, we chose the "Instructions" part of the Alpaca-52k dataset as list elements. In addition, to ensure the appropriateness of text length, all selected instructions were limited to 5\textasciitilde40 tokens.

\subsection{MultiDoc}
\label{appdix:multidoc detail}
To construct contexts $X$ in this scenario, we selected documents from four datasets: GovReport~\cite{huang2021efficientgovreport}, XSum~\cite{narayan2018donxsum}, QMSum~\cite{zhong2021qmsum}, and the Paul Graham Essays\footnote{\url{https://huggingface.co/datasets/sgoel9/paul_graham_essays}}. 
 GovReport is a long document summary dataset consisting of reports written by government research agencies; XSum and QMSum are respectively a news summary dataset and a metting summary dataset , both of which cover a wide range of domains; and the Paul Graham Essays dataset collects articles written by Paul Graham in a variety of fields. These selective datasets span multiple domains and various forms of text, offering a high degree of diversity. 

For each dataset, we extract the main text (excluding summaries) to construct multi-document task contexts $X$, limiting each text to 300–500 tokens through filtering and truncation.
As shown in Table \ref{tab:allprompts}, each document contains six fields: "text", "id", "iD2", "title", "date", and "source". The "text" field holds the processed content, while the other five are attributes designed for the tasks, partially sourced from the dataset and partially constructed manually. Each document has unique "id" and "iD2" fields, though some may lack "title" or "source". To reduce LLMs' reliance on parameter knowledge, we randomly annotate the "source" field, breaking its correlation with the "text". This ensures the model cannot infer the source using pre-trained knowledge, creating additional challenges for instruction adherence. Lastly, we introduced duplicates by reusing some "text" fields in different documents, maintaining a duplication rate of 25\%. 

\subsection{OneDoc}
To create the context $X$ for OneDoc, we synthesized an extra-long document by concatenating entries from the Paul Graham Essays dataset, following the approach in \cite{gkamradt2023llmtest}. Additionally, some sentences are randomly tagged as key information, with each tag specifying a type (e.g., Topic, Evidence, Concession) and a unique identifier to aid LLMs in identifying and categorizing critical content.

\section{Effectiveness of Expression Extension}
\label{appdix: effective EE}

\subsectiontitle{Implementation details} To implement RECS for expression extension, we chose the BGE-M3 text embedding model \cite{chen2024bge} for encoding and applied K-means clustering \cite{macqueen1967kmeans}. Additionally, since LLMs can introduce inaccuracies in rewrites due to misinterpretations, we manually filtered out unsuitable rewrites during the sampling phase, resulting in 5\textasciitilde6 instruction templates for each task.

\begin{figure}[ht]
\small
\begin{tcolorbox}[promptbox, title=Prompt for Rewriting in RECS]
\small
Please rewrite the given prompt according to the following requirements:

    1. The rewritten prompt must retain the same meaning as the original, without altering the intent.\\
    2. Try your best to use different vocabulary or sentence structures.\\
    3. Ensure that the rewritten prompt is clear and accurate, avoiding any expressions that could lead to ambiguity.\\
    4. Please keep the placeholders in the prompt (i.e., “\{\}” and the contents therein) exactly as they are during rewriting.\\
    5. Please keep the example in the prompt, but you can make some small changes while keeping the original meaning.\\
    6. Output the result in Json List format, without anything else.\\
    7. Please generate 20 different rewrites at once. 
    
prompt: \textit{\textbf{\{Prompt to be rewritten\}}}
\end{tcolorbox}
\vspace{-2.5ex}
\caption{Prompt template for rewriting task prompt.}
\vspace{-4ex}
\end{figure}

\subsectiontitle{Validity Experiment} To validate the effectiveness of the RECS method proposed in Section \ref{sec: Data Extension}, we compared it with a random sampling approach. Specifically, after the Rewriting phase, we constructed a dataset by randomly sampling (rather than clustering) the same number of instructions and calculated the IFS values for sixteen models on this dataset. 

As shown in \ref{fig:gap}, compared to random sampling, RECS led to higher expression IFS scores in 75\% of the models, outperforming the IFS improvement rates observed for the Variable and Length perspectives, at 37.5\% and 62.5\% respectively. Additionally, when comparing the mean IFS values across all models, we observed that RECS improved expression IFS while slightly lowering variable IFS and length IFS. This suggests that the RECS method make it more challenging for models to maintain stability when handling varied expressions, which shows the effectiveness in expression extension as well.

\section{Sampling Space for Instruction Variables}
\label{app:instruction-variable}
\begin{figure}[ht]
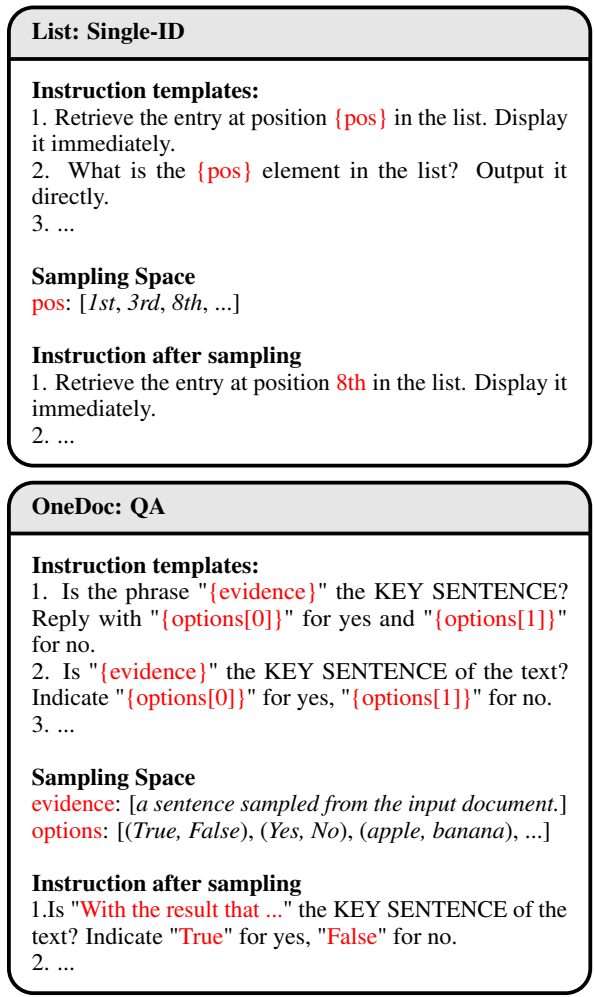

\small
\begin{tcolorbox}[promptbox, title=List: Single-ID]
\small
\textbf{Instruction templates:} \\
1. Retrieve the entry at position \textcolor{red}{\{pos\}} in the list. Display it immediately.\\
2. What is the \textcolor{red}{\{pos\}} element in the list? Output it directly.\\
3. ...\\

\textbf{Sampling Space}\\
\textcolor{red}{pos}: [\textit{1st}, \textit{3rd}, \textit{8th}, ...] \\

\textbf{Instruction after sampling}\\
1. Retrieve the entry at position \textcolor{red}{8th} in the list. Display it immediately.\\
2. ...
\end{tcolorbox}

\begin{tcolorbox}[promptbox, title=OneDoc: QA]
\small
\textbf{Instruction templates:}\\
1. Is the phrase "\textcolor{red}{\{evidence\}}" the KEY SENTENCE? Reply with "\textcolor{red}{\{options[0]\}}" for yes and "\textcolor{red}{\{options[1]\}}" for no.\\
2. Is "\textcolor{red}{\{evidence\}}" the KEY SENTENCE of the text? Indicate "\textcolor{red}{\{options[0]\}}" for yes, "\textcolor{red}{\{options[1]\}}" for no.\\
3. ...\\

\textbf{Sampling Space}\\
\textcolor{red}{evidence}: [\textit{a sentence sampled from the input document.}]\\
\textcolor{red}{options}: [(\textit{True, False}), (\textit{Yes, No}), (\textit{apple, banana}), ...]\\
 
\textbf{Instruction after sampling}\\
1.Is "\textcolor{red}{With the result that ...}" the KEY SENTENCE of the text? Indicate "\textcolor{red}{True}" for yes, "\textcolor{red}{False}" for no.\\
2. ...
\end{tcolorbox}
\vspace{-2ex}
\caption{Examples of the sampling space. \textcolor{red}{Red} indicates the instruction variables $var$, and \{\} indicates a placeholder in the instruction.}
\label{fig:variable-example}
\vspace{-2ex}
\end{figure}

As shown in Figure \ref{fig:variable-example}, we reserve identical placeholders for each instruction template and set the sampling space accordingly. Depending on the task requirements, the types of instruction variables include numerical values, lists, phrases, long sentences, and format indicators, etc. We carefully consider variable distribution to avoid biases from the sampling mechanism. For example, in position-related variables within the List scenario, inputs were divided into three sections (beginning, middle, and end), with the middle section receiving the largest portion. We then randomly sample 2–3 elements from each section to ensure balance. Additionally, we introduce unconventional causal relationships in certain rule-based variables to increase task difficulty. For instance, in the QA task, models are required to use "False" for correct answers and "True" for incorrect ones, making it more challenging to follow instructions.

\section{Rubric~Design~and~Evaluation~Program}
\label{app:rubric design}
\subsectiontitle{Rubric Design}
For the weight \(\tilde s\) assignment of scoring points, we primarily considered three factors:

(1) \textbf{Evaluation complexity}: Scoring points with higher evaluation complexity require more assessment steps and are therefore assigned greater weight. For example, in assessing format correctness, the Batch-Label (MB) task is given higher scoring weights compared to the QA (OQ) task due to its more complex formatting requirements.

(2) \textbf{Task difficulty for models}: Some scoring points may not be as complex to evaluate but are particularly challenging for the model to fulfill. In such cases, we allocate greater weight to these points as well.

(3) \textbf{Balance across capabilities and scenarios}: Based on the considerations in (1) and (2), we made fine adjustments to ensure that the weights assigned to different instruction-following capabilities and scenarios are as balanced as possible. This balance is shown in Figure \ref{fig:stastic}.

On this basis, we present the rubric of \ourseval in Table \ref{tab:rubric}. Additionally, the pseudocode examples reflect our considerations above. Program \ref{alg:json} show the correspondence between evaluation steps and their respective weights. In Program \ref{alg:quantity}, we award an additional point for responses that fully meet the requirements, highlighting the difficulty of that particular scoring point.

\subsectiontitle{Evaluation Program} As described in Section \ref{sec:ars}, we design an automated evaluation program based on the scoring rubric to evaluate the quality of LLM responses. Inspired by \cite{ifezhou2023instruction}, during task design, we aim to ensure that the correct answers could be captured by the program. However, our tasks incorporate more complex formatting and logical constraints, making it challenging to directly obtain accurate evaluation results through simple automated methods. To address this, we iteratively decomposed and refined each scoring point $s$ into sub-evaluation criteria that could be directly assessed by the program, enabling a more detailed and discriminative evaluation.

Our program evaluation adheres to two principles: \textbf{(1) correct answers should achieve full scores, and (2) partially correct answers should be distinguishable in terms of scores.} In general, we first extract structured information from the outputs based on the format constraints of each task, and then perform evaluations on other dimensions using this information. However, LLM responses are not always well-structured and may contain minor errors that hinder the extraction of structured information. To mitigate this, we carefully designed task-specific validation programs that employ techniques such as regex matching and substring retrieval to maximize the extraction of valid part from LLM responses (see Program \ref{alg:json}). 
Finally, we manually provided reference answers for all samples in \ours. We verified that \textbf{the automated evaluation programs in \ourseval consistently achieve full scores on all reference answers}, thereby fulfilling the first principle outlined above.

\section{Experiments Setup}
\label{app:experimentsetup}
\subsectiontitle{Baselines} For closed-source models, we specifically use gpt-4o-2024-08-06 and gpt-4-0125-preview to represent the performance of GPT-4o and GPT-4, respectively. For open-source models, all implementations were sourced from Hugging~Face\footnote{\url{https://huggingface.co/}} (as detailed in Table 5), and we test the performance of both base models and fine-tuned models. Models with the suffix "-Inst." or "-chat" indicate that they have been fine-tuned on instruction or dialogue data. 

\subsectiontitle{Inference} During the inference process, we complete the deployment of all open-source models with the vLLM \cite{kwon2023efficient}. The temperature is set to 0 to ensure deterministic outputs. For the SFT versions of the models, we used the official chat template. For the base versions, we add a suffix "Output: " to prompt the models to generate answers according to the instructions. Token counts were calculated using GPT-4's tokenizer\footnote{\url{https://github.com/openai/tiktoken}}, and truncation was applied to adjust context for models unable to process the longest contexts. To ensure diversity and accuracy in instruction phrasing, we generated 5–6 template variants for each original instruction and manually filtered them. Each template included placeholders for variables (e.g., numerical values, lists, phrases, long sentences, and format indicators), dynamically sampled from curated sets of 5–10 candidates to create task-specific prompts. In order to avoid unnecessary time consumption resulting from endless repetitions in the output, we set maximum generation lengths for different tasks: List scenario tasks that output a single element were limited to 100 tokens, tasks in the MultiDoc scenario are set to 4096 tokens, and the rest of the tasks are limited to 512 tokens.

Due to variations in encoding efficiency across models' tokenizer, some models can not accommodate the longest inputs. To address this, we apply right truncation to the context $X$, ensuring the completeness of scenario description~$D$ and instruction~$I$. The experiments were conducted across six context length intervals, ranging from 4k to 128k tokens, with task-specific output limits to ensure sufficient space for model generation. 

\begin{table}[t]
  \centering
  \resizebox{.99\columnwidth}{!}{
  \begin{tabular}{l|c}
  \rowcolor[gray]{0.9}
    \Xhline{1pt}
    \textbf{Model} & \textbf{Hugging Face ID} \\ \hline
    Llama-3.1-70B-Inst. & meta-llama/Llama-3.1-70B-Instruct \\
    Llama-3.1-8B-Inst. &  meta-llama/Llama-3.1-8B-Instruct \\
    Llama-3.1-70B &  meta-llama/Llama-3.1-70B \\
    Llama-3.1-8B & meta-llama/Llama-3.1-8B \\ \hdashline
    Qwen2.5-72B-Inst. & Qwen/Qwen2.5-72B-Instruct\\
    Qwen2.5-32B-Inst. & Qwen/Qwen2.5-72B-Instruct\\
    Qwen2.5-14B-Inst. & Qwen/Qwen2.5-72B-Instruct\\
    Qwen2.5-7B-Inst. & Qwen/Qwen2.5-72B-Instruct\\
    Qwen2.5-72B & Qwen/Qwen2.5-72B\\
    Qwen2.5-32B & Qwen/Qwen2.5-72B\\
    Qwen2.5-14B & Qwen/Qwen2.5-72B\\
    Qwen2.5-7B & Qwen/Qwen2.5-72B \\ \hdashline
    C4AI-cmd-r-08-2024 & CohereLabs/c4ai-command-r-08-2024 \\
    C4AI-cmd-r-v01 & CohereLabs/c4ai-command-r-v01 \\ \hdashline
    LWM-Text-Chat-1M  & LargeWorldModel/LWM-Text-Chat-1M \\
    LWM-Text-1M & LargeWorldModel/LWM-Text-1M \\ \hdashline
    InternLM2.5-7b-chat-1m & internlm/internlm2\_5-7b-chat-1m \\ \hdashline
    GLM-4-9b-chat-1m & THUDM/glm-4-9b-chat-1m\\
    
    \Xhline{1pt}
  \end{tabular}}
 \vspace{-1ex}
  \caption{A mapping between the Hugging Face model IDs and the aliases used in the paper.}
  \vspace{-2ex}
  \label{tab:huggingface-map}

\end{table}

\raggedbottom  

\section{Inspiration for IFS}
IFS is derived from the coefficient of variation (CV)~\cite{brown1998cv}, a statistical measure widely used in fields such as risk assessment and quality control~\cite{shechtman2013coefficient}. CV is calculated as the ratio of the standard deviation to the mean, providing a normalized measure of variability.

Using standard deviation alone to measure instruction-following stability can lead to misleading conclusions. For instance, if a model's performance across all intervals is zero, its standard deviation would also be zero, suggesting perfect stability despite the model's complete inability to perform. IFS avoids this issue by normalizing variability relative to the mean, ensuring that stability is assessed in a scale-independent and interpretable manner.

The application of CV to instruction-following tasks aligns well with its traditional use in evaluating consistency and reliability across varying scenarios. By capturing fluctuations in model performance relative to its average capability, IFS offers a robust and fair metric for comparing the stability of different LLMs.
\vspace{2ex}

\begin{algorithm}[htbp]
\small
\caption{Format Validation (JSON Dict)}\label{alg:json}
\textbf{Target}: Verify if the output meets the JSON Dict format.
\begin{algorithmic}[1]
\Require LLM answer $a$, full score $\tilde{s}=4$
\Ensure Validation score $s_a=f_s(a)\in [0, \tilde{s}]$
\vspace{1ex}
\Statex \textbf{Step 1: Symbol Check (Max 1 point)}
\State $s_a \gets 0$
\If{$\text{count}(\texttt{\{\}}) \geq 2 \land \text{count}(\texttt{"}) \geq 4 
\land \text{count}(\texttt{:}) \geq 1$} 
    \State $s_a\gets s_a+ 1$ 
\EndIf
\vspace{1ex}
\Statex \textbf{Step 2: Parsing Check (Max 2 points)}
\State $a' \gets \text{regex\_extract}(a, \texttt{'\textbackslash\{[\^{}\textbackslash]]+\textbackslash\}'})$ 
\If{$\text{json.loads}(a)$ succeeds} 
    \State $s_a\gets s_a+ 2$
\ElsIf{$\text{json.loads}(a')$ succeeds} 
    \State $s_a\gets s_a+ 1$
\EndIf
\vspace{1ex}
\Statex \textbf{Step 3: KV Check (Max 1 point)}
\If{$\text{check\_key\_value\_format}(a / a')$ is correct} 
    \State $s_a\gets s_a+ 1$ 
\EndIf

\State \Return $s_a$ 
\end{algorithmic}
\vspace{0.5em} 
\footnotesize 
\textbf{Note:} The function $\text{check\_key\_value\_format}(\cdot)$ serves as an example for additional format checks, with specifics depending on the task requirements.
\end{algorithm}

\begin{algorithm}[H]
\small
\caption{Quantity Verification (LMI)}\label{alg:quantity}
\textbf{Target}: Verify if the number of output elements is correct.
\begin{algorithmic}[1]
\Require LLM answer $a$, target number $n_{\text{gold}}$, full score $\tilde{s} = 3$
\Ensure Validation score $s_a = f_s(a) \in [0, \tilde{s}]$

\State $n_{\text{pred}} \gets \text{extract\_quantity}(a)$ 
\Statex \Comment{Extract the number of elements in the output}

\If{$n_{\text{pred}} = n_{\text{gold}}$}
    \State $s_a \gets \tilde{s}$ \Comment{Full score if the quantity matches}
\Else
    \State $s_a \gets \max\left(0, \left(1.0 - \frac{|n_{\text{pred}} - n_{\text{gold}}|}{n_{\text{gold}}} \right) \right) \times 2$ 
\Statex\Comment{Penalty based on relative deviation}
\EndIf

\State \Return $s_a$
\end{algorithmic}
\textbf{Note:} The function \( n_{\text{pred}} = \text{extract\_quantity}(a) \) can be executed in various ways. If the LLM answer \( a \) passes format checks and is parsable, the target quantity is directly obtained. Otherwise, string processing techniques are employed to extract it with feedback.
\end{algorithm}

\begin{table*}[!ht]
  \centering
  \renewcommand{\arraystretch}{1.3}
  \resizebox{\textwidth}{!}{
  \begin{tabular}{c|ccc}
    \Xhline{1pt}
    \rowcolor[gray]{0.9}
    \textbf{Tasks} & \textbf{Score Point~$s$} & \textbf{Weight~$\tilde{s}$} & \textbf{Related Capability} \\
    \hline
    ~ & Format correctness. & 1 & \textit{Fmt} \\
    ~ & Answer is from the input list. & 2 & \textit{Ori} \\
    ~ & Answer correctness. & 1 & \textit{Recog} \\ \cdashline{2-4} \rowcolor{gray!10}
     \cellcolor{white}{\multirow{-4}{*}{LSI, LOI, LOE}} & {\textit{Total Weight}~($\tilde R$)} & 4 & -\\ \hline
    ~ & Format correctness. & 2 & \textit{Fmt} \\
    ~ & Order correctness. & 2 & \textit{Spat} \\ 
    ~ & The number of output elements is correct. & 3 & \textit{Num} \\
    ~ & Answers correctness. & 3 & \textit{Ori} \\\cdashline{2-4} \rowcolor{gray!10}
    \cellcolor{white}{\multirow{-5}{*}{LMI}} & {\textit{Total Weight}~($\tilde{R}$)} & 10 & -\\ \hline
    ~ & Format correctness. & 1 & \textit{Fmt} \\
    ~ & Answer is from the input list. & 1 & \textit{Ori} \\
    ~ & The output element conform to the position constraint. & 3 & \textit{Spat} \\ \cdashline{2-4} \rowcolor{gray!10}
    \cellcolor{white}{\multirow{-4}{*}{LBI, LBE}} & {\textit{Total Weight}~($\tilde{R}$)} & 5 & -\\ \hline
    ~ & Answer correctness (conforms to label logic). & 3 & \textit{Logit} \\
    ~ & Output labels are from the candidate set. & 3 & \textit{Ori} \\
    ~ & The number of output labels matches the number of input documents. & 3 & \textit{Num, Recog} \\
    ~ & Format correctness. & 5 & \textit{Fmt} \\ \cdashline{2-4} \rowcolor{gray!10}
    \cellcolor{white}{\multirow{-5}{*}{MB}} & {\textit{Total Weight}~($\tilde{R}$)} & 14 & -\\ \hline
    ~ & Answer correctness. & 4 & \textit{Logit, Recog} \\
    ~ & Find the correct number of duplicate documents. & 5 & \textit{Num, Logit} \\
    ~ & Format correctness. & 5 & \textit{Fmt} \\ 
    ~ & Document properties in the output are in the input & 6 & \textit{Ori} \\ \cdashline{2-4} \rowcolor{gray!10}
    \cellcolor{white}{\multirow{-5}{*}{MF}} & {\textit{Total Weight}~($\tilde{R}$)} & 20 & -\\ \hline
    ~ & Answer correctness. & 3 & \textit{Logit} \\
    ~ & Output sentences are from the input document. & 2 & \textit{Ori} \\
    ~ & Format correctness. & 3 & \textit{Fmt} \\ 
    ~ & Output sentences are the key sentence. & 2 & \textit{Recog} \\ 
    ~ & The number of output key sentences is correct. & 4 & \textit{Num} \\ \cdashline{2-4} \rowcolor{gray!10}
    \cellcolor{white}{\multirow{-6}{*}{OR}} & {\textit{Total Weight}~($\tilde{R}$)} & 14 & -\\ \hline
    ~ & Answer correctness. & 3 & \textit{Logit} \\
    ~ & Format correctness. & 2 & \textit{Fmt} \\ \cdashline{2-4} \rowcolor{gray!10}
    \cellcolor{white}{\multirow{-3}{*}{OQ}} & {\textit{Total Weight}~($\tilde{R}$)} & 5 & -\\ \hline
    ~ & Output sentences are from the input document. & 2 & \textit{Ori} \\
    ~ & Format correctness. & 4 & \textit{Fmt} \\
    ~ & Output sentences are target key sentences. & 4 & \textit{Recog} \\ 
    ~ & The output order matches the ids sort. & 4 & \textit{Spat} \\ \cdashline{2-4} \rowcolor{gray!10}
    \cellcolor{white}{\multirow{-4}{*}{OE}} & {\textit{Total Weight}~($\tilde{R}$)} & 14 & -\\ \hline
    \Xhline{1pt}
  \end{tabular}}
  \caption{The scoring rubric and Score-Capability Map in \ourseval.}
  \label{tab:rubric}
\end{table*}


\clearpage
\onecolumn
\section{Full Results on \ours}

\subsection{ARS Score}
\begin{table}[!h]
    \small
    \rowcolors{2}{white}{gray!10}
    \resizebox{\textwidth}{!}{
    \begin{tabular}{c|ccccp{1pt}ccccccp{1pt}ccp{1pt}c}
        \arrayrulecolor{black}
           \Xhline{1pt}
           \rowcolor[gray]{0.9} 
           & & \multicolumn{3}{c}{\textbf{OneDoc}} & & \multicolumn{6}{c}{\textbf{List}} & & \multicolumn{2}{c}{\textbf{MultiDoc}} & &\\
           \noalign{\vskip -2pt} \cmidrule(lr){3-5} \cmidrule(lr){7-12}  \cmidrule(lr){14-15} 
            \rowcolor[gray]{0.9} \noalign{\vskip -3pt}
           \multirow{-2}{*}{\textbf{Model}}& \multirow{-2}{*}{\textbf{Length}} &\textbf{OR} & \textbf{OQ} & \textbf{OE} & & \textbf{LSI} & \textbf{LMI} & \textbf{LOI} & \textbf{LOE} & \textbf{LBI} & \textbf{LBE} & & \textbf{MB} & \textbf{MF} & &\multirow{-2}{*}{\textbf{Overall}}\\  \hline
           
           \cellcolor{white} ~ & 4k & 0.770  & 0.831  & 0.781  & ~ & 0.942  & 0.938  & 0.835  & 0.874  & 0.755  & 0.866  & ~ & 0.735  & 0.627  & ~ & 0.776  \\
           \cellcolor{white} ~ & 8k & 0.859  & 0.863  & 0.893  & ~ & 0.942  & 0.946  & 0.707  & 0.856  & 0.754  & 0.830  & ~ & 0.782  & 0.637  & ~ & 0.807   \\ 
           \cellcolor{white} ~ & 16k & 0.830  & 0.863  & 0.812  & ~ & 0.925  & 0.885  & 0.778  & 0.849  & 0.794  & 0.856  & ~ & 0.749  & 0.710  & ~ & 0.801   \\ 
           \cellcolor{white} ~ & 32k & 0.799  & 0.901  & 0.831  & ~ & 0.863  & 0.871  & 0.745  & 0.808  & 0.767  & 0.871  & ~ & 0.778  & 0.617  & ~ & 0.779   \\ 
           \cellcolor{white} ~ & 64k & 0.758  & 0.935  & 0.829  & ~ & 0.829  & 0.755  & 0.773  & 0.826  & 0.728  & 0.753  & ~ & 0.746  & 0.468  & ~ & 0.721   \\ 
           \multirow{-6}{*}{\cellcolor{white}GPT-4o} & 128k & 0.766  & 0.895  & 0.857  & ~ & 0.783  & 0.619  & 0.601  & 0.726  & 0.699  & 0.775  & ~ & 0.526  & 0.468  & ~ & 0.666   \\ \hline
           \cellcolor{white} ~ & 4k & 0.939  & 0.832  & 0.907  & ~ & 0.875  & 0.923  & 0.789  & 0.819  & 0.790  & 0.896  & ~ & 0.702  & 0.879  & ~ & 0.859   \\ 
           \cellcolor{white} ~ & 8k & 0.953  & 0.865  & 0.926  & ~ & 0.954  & 0.871  & 0.795  & 0.811  & 0.790  & 0.869  & ~ & 0.715  & 0.836  & ~ & 0.854   \\ 
           \cellcolor{white} ~ & 16k & 0.777  & 0.793  & 0.802  & ~ & 0.954  & 0.801  & 0.769  & 0.824  & 0.731  & 0.885  & ~ & 0.744  & 0.843  & ~ & 0.804   \\ 
           \cellcolor{white} ~ & 32k & 0.527  & 0.787  & 0.750  & ~ & 0.975  & 0.684  & 0.715  & 0.807  & 0.746  & 0.815  & ~ & 0.552  & 0.756  & ~ & 0.700   \\ 
           \cellcolor{white} ~ & 64k & 0.526  & 0.819  & 0.537  & ~ & 0.833  & 0.584  & 0.648  & 0.761  & 0.687  & 0.808  & ~ & 0.681  & 0.663  & ~ & 0.647   \\ 
           \multirow{-6}{*}{GPT-4} & 128k & 0.524  & 0.825  & 0.494  & ~ & 0.767  & 0.547  & 0.510  & 0.665  & 0.757  & 0.719  & ~ & 0.207  & 0.685  & ~ & 0.561   \\ \hline
           \cellcolor{white} ~ & 4k  & 0.823  & 0.773  & 0.907  & ~ & 0.954  & 0.853  & 0.767  & 0.813  & 0.820  & 0.867  & ~ & 0.698  & 0.876  & ~ & 0.833   \\
           \cellcolor{white} ~ & 8k  & 0.840  & 0.781  & 0.892  & ~ & 0.942  & 0.782  & 0.774  & 0.731  & 0.858  & 0.846  & ~ & 0.697  & 0.783  & ~ & 0.805   \\ 
           \cellcolor{white} ~ & 16k  & 0.722  & 0.727  & 0.822  & ~ & 0.900  & 0.630  & 0.703  & 0.682  & 0.837  & 0.781  & ~ & 0.704  & 0.738  & ~ & 0.741   \\ 
           \cellcolor{white} ~ & 32k  & 0.768  & 0.731  & 0.803  & ~ & 0.842  & 0.615  & 0.714  & 0.741  & 0.876  & 0.841  & ~ & 0.684  & 0.557  & ~ & 0.710   \\ 
           \cellcolor{white} ~ & 64k  & 0.814  & 0.874  & 0.685  & ~ & 0.850  & 0.589  & 0.583  & 0.594  & 0.755  & 0.812  & ~ & 0.713  & 0.332  & ~ & 0.645   \\ 
           \multirow{-6}{*}{Qwen2.5-72B-Inst.\textsuperscript{$\dagger$}} & 128k  & 0.583  & 0.759  & 0.789  & ~ & 0.717  & 0.574  & 0.535  & 0.524  & 0.765  & 0.693  & ~ & 0.157  & 0.220  & ~ & 0.502   \\ \hline
           \cellcolor{white} ~ & 4k  & 0.952  & 0.838  & 0.941  & ~ & 0.983  & 0.898  & 0.803  & 0.863  & 0.842  & 0.831  & ~ & 0.555  & 0.709  & ~ & 0.814   \\ 
           \cellcolor{white} ~ & 8k  & 0.884  & 0.864  & 0.943  & ~ & 0.979  & 0.948  & 0.809  & 0.919  & 0.800  & 0.884  & ~ & 0.765  & 0.702  & ~ & 0.842   \\ 
           \cellcolor{white} ~ & 16k  & 0.903  & 0.806  & 0.894  & ~ & 0.892  & 0.884  & 0.786  & 0.858  & 0.802  & 0.841  & ~ & 0.778  & 0.617  & ~ & 0.804   \\ 
           \cellcolor{white} ~ & 32k  & 0.560  & 0.843  & 0.829  & ~ & 0.875  & 0.897  & 0.680  & 0.809  & 0.726  & 0.813  & ~ & 0.630  & 0.568  & ~ & 0.707   \\ 
           \cellcolor{white} ~ & 64k  & 0.609  & 0.928  & 0.624  & ~ & 0.742  & 0.761  & 0.525  & 0.770  & 0.715  & 0.831  & ~ & 0.772  & 0.381  & ~ & 0.645   \\ 
           \multirow{-6}{*}{Llama-3.1-70B-Inst.}~ & 128k  & 0.475  & 0.882  & 0.036  & ~ & 0.358  & 0.402  & 0.302  & 0.571  & 0.270  & 0.531  & ~ & 0.444  & 0.209  & ~ & 0.353   \\ \hline
           \cellcolor{white} ~ & 4k  & 0.920  & 0.767  & 0.897  & ~ & 0.736  & 0.706  & 0.626  & 0.587  & 0.789  & 0.826  & ~ & 0.840  & 0.800  & ~ & 0.808   \\ 
           \cellcolor{white} ~ & 8k  & 0.728  & 0.855  & 0.919  & ~ & 0.692  & 0.719  & 0.560  & 0.548  & 0.840  & 0.796  & ~ & 0.826  & 0.633  & ~ & 0.749   \\ 
           \cellcolor{white} ~ & 16k  & 0.670  & 0.754  & 0.907  & ~ & 0.721  & 0.574  & 0.565  & 0.547  & 0.789  & 0.756  & ~ & 0.711  & 0.529  & ~ & 0.678   \\ 
           \cellcolor{white} ~ & 32k  & 0.736  & 0.815  & 0.858  & ~ & 0.683  & 0.596  & 0.572  & 0.602  & 0.748  & 0.719  & ~ & 0.146  & 0.402  & ~ & 0.578   \\ 
           \cellcolor{white} ~ & 64k  & 0.598  & 0.809  & 0.829  & ~ & 0.646  & 0.540  & 0.549  & 0.475  & 0.731  & 0.780  & ~ & 0.742  & 0.255  & ~ & 0.597   \\ 
           \multirow{-6}{*}{Qwen2.5-32B-Inst.\textsuperscript{$\dagger$}} & 128k  & 0.557  & 0.750  & 0.687  & ~ & 0.494  & 0.540  & 0.380  & 0.341  & 0.682  & 0.689  & ~ & 0.386  & 0.238  & ~ & 0.489   \\ \hline
           \cellcolor{white} ~ & 4k & 0.620  & 0.849  & 0.819  & ~ & 0.783  & 0.767  & 0.738  & 0.718  & 0.750  & 0.809  & ~ & 0.773  & 0.936  & ~ & 0.791   \\ 
           \cellcolor{white} ~ & 8k & 0.689  & 0.824  & 0.682  & ~ & 0.742  & 0.758  & 0.569  & 0.659  & 0.744  & 0.838  & ~ & 0.827  & 0.824  & ~ & 0.755   \\ 
           \cellcolor{white} ~ & 16k & 0.609  & 0.869  & 0.573  & ~ & 0.858  & 0.588  & 0.420  & 0.531  & 0.646  & 0.484  & ~ & 0.786  & 0.752  & ~ & 0.664   \\ 
           \cellcolor{white} ~ & 32k & 0.499  & 0.895  & 0.546  & ~ & 0.763  & 0.570  & 0.471  & 0.564  & 0.648  & 0.531  & ~ & 0.803  & 0.592  & ~ & 0.616   \\ 
           \cellcolor{white} ~ & 64k & 0.391  & 0.769  & 0.360  & ~ & 0.608  & 0.477  & 0.401  & 0.400  & 0.602  & 0.500  & ~ & 0.809  & 0.454  & ~ & 0.512   \\ 
           \multirow{-6}{*}{C4AI-cmd-r-08-2024} & 128k & 0.364  & 0.822  & 0.228  & ~ & 0.400  & 0.556  & 0.366  & 0.308  & 0.612  & 0.530  & ~ & 0.373  & 0.402  & ~ & 0.416   \\ \hline
           \cellcolor{white} ~ & 4k & 0.664  & 0.798  & 0.559  & ~ & 0.746  & 0.650  & 0.536  & 0.505  & 0.692  & 0.604  & ~ & 0.799  & 0.916  & ~ & 0.715   \\ 
           \cellcolor{white} ~ & 8k & 0.570  & 0.863  & 0.450  & ~ & 0.767  & 0.668  & 0.520  & 0.502  & 0.685  & 0.566  & ~ & 0.765  & 0.794  & ~ & 0.659   \\ 
           \cellcolor{white} ~ & 16k & 0.496  & 0.797  & 0.386  & ~ & 0.567  & 0.573  & 0.528  & 0.470  & 0.637  & 0.598  & ~ & 0.711  & 0.636  & ~ & 0.577   \\ 
           \cellcolor{white} ~ & 32k & 0.461  & 0.842  & 0.448  & ~ & 0.588  & 0.555  & 0.442  & 0.558  & 0.690  & 0.676  & ~ & 0.832  & 0.593  & ~ & 0.598   \\ 
           \cellcolor{white} ~ & 64k & 0.360  & 0.780  & 0.356  & ~ & 0.663  & 0.512  & 0.495  & 0.574  & 0.731  & 0.780  & ~ & 0.763  & 0.467  & ~ & 0.541   \\ 
           \multirow{-6}{*}{C4AI-cmd-r-v01} & 128k & 0.417  & 0.829  & 0.321  & ~ & 0.517  & 0.515  & 0.414  & 0.429  & 0.729  & 0.633  & ~ & 0.458  & 0.471  & ~ & 0.482   \\ \hline
           \cellcolor{white} ~ & 4k & 0.576  & 0.738  & 0.674  & ~ & 0.709  & 0.734  & 0.560  & 0.604  & 0.697  & 0.779  & ~ & 0.815  & 0.834  & ~ & 0.722   \\ 
           \cellcolor{white} ~ & 8k & 0.505  & 0.753  & 0.598  & ~ & 0.775  & 0.611  & 0.516  & 0.503  & 0.671  & 0.726  & ~ & 0.765  & 0.691  & ~ & 0.647   \\ 
           \cellcolor{white} ~ & 16k & 0.360  & 0.705  & 0.500  & ~ & 0.592  & 0.641  & 0.529  & 0.504  & 0.681  & 0.648  & ~ & 0.760  & 0.500  & ~ & 0.563   \\ 
           \cellcolor{white} ~ & 32k & 0.294  & 0.647  & 0.484  & ~ & 0.425  & 0.504  & 0.426  & 0.504  & 0.663  & 0.728  & ~ & 0.839  & 0.501  & ~ & 0.538   \\ 
           \cellcolor{white} ~ & 64k & 0.321  & 0.596  & 0.073  & ~ & 0.508  & 0.462  & 0.362  & 0.523  & 0.597  & 0.598  & ~ & 0.891  & 0.291  & ~ & 0.434   \\ 
           \multirow{-6}{*}{Qwen2.5-72B\textsuperscript{$\dagger$}} & 128k & 0.355  & 0.725  & 0.243  & ~ & 0.613  & 0.522  & 0.323  & 0.345  & 0.542  & 0.517  & ~ & 0.288  & 0.314  & ~ & 0.383   \\ \hline
           \cellcolor{white} ~ & 4k & 0.889  & 0.187  & 0.095  & ~ & 0.767  & 0.635  & 0.686  & 0.631  & 0.778  & 0.745  & ~ & 0.431  & 0.295  & ~ & 0.494   \\ 
           \cellcolor{white} ~ & 8k & 0.091  & 0.264  & 0.047  & ~ & 0.633  & 0.423  & 0.530  & 0.553  & 0.606  & 0.694  & ~ & 0.915  & 0.367  & ~ & 0.414   \\ 
           \cellcolor{white} ~ & 16k & 0.441  & 0.290  & 0.179  & ~ & 0.783  & 0.446  & 0.547  & 0.509  & 0.721  & 0.644  & ~ & 0.892  & 0.221  & ~ & 0.461   \\ 
           \cellcolor{white} ~ & 3k2 & 0.117  & 0.300  & 0.330  & ~ & 0.713  & 0.361  & 0.598  & 0.686  & 0.715  & 0.728  & ~ & 0.899  & 0.308  & ~ & 0.458   \\ 
           \cellcolor{white} ~ & 64k & 0.218  & 0.300  & 0.021  & ~ & 0.742  & 0.282  & 0.603  & 0.713  & 0.716  & 0.729  & ~ & 0.852  & 0.179  & ~ & 0.390   \\ 
           \multirow{-6}{*}{Llama-3.1-70B} & 128k & 0.264  & 0.300  & 0.036  & ~ & 0.370  & 0.219  & 0.190  & 0.394  & 0.633  & 0.547  & ~ & 0.261  & 0.479  & ~ & 0.311   \\ \hline
           \cellcolor{white} ~ & 4k & 0.515  & 0.809  & 0.553  & ~ & 0.586  & 0.554  & 0.487  & 0.570  & 0.586  & 0.579  & ~ & 0.792  & 0.508  & ~ & 0.588   \\ 
           \cellcolor{white} ~ & 8k & 0.582  & 0.813  & 0.483  & ~ & 0.561  & 0.585  & 0.444  & 0.530  & 0.687  & 0.651  & ~ & 0.757  & 0.567  & ~ & 0.602   \\ 
           \cellcolor{white} ~ & 16k & 0.338  & 0.817  & 0.429  & ~ & 0.055  & 0.306  & 0.187  & 0.368  & 0.254  & 0.221  & ~ & 0.859  & 0.307  & ~ & 0.413   \\ 
           \cellcolor{white} ~ & 32k & 0.296  & 0.503  & 0.144  & ~ & 0.033  & 0.100  & 0.005  & 0.000  & 0.052  & 0.000  & ~ & 0.926  & 0.397  & ~ & 0.313   \\ 
           \cellcolor{white} ~ & 64k & 0.128  & 0.133  & 0.036  & ~ & 0.297  & 0.358  & 0.223  & 0.250  & 0.255  & 0.196  & ~ & 0.907  & 0.310  & ~ & 0.311   \\ 
           \multirow{-6}{*}{Qwen2.5-32B\textsuperscript{$\dagger$}} & 128k & 0.223  & 0.100  & 0.036  & ~ & 0.050  & 0.100  & 0.050  & 0.117  & 0.195  & 0.137  & ~ & 0.150  & 0.191  & ~ & 0.137   \\ \hline
           \cellcolor{white} ~ & 4k & 0.843  & 0.761  & 0.731  & ~ & 0.697  & 0.740  & 0.480  & 0.365  & 0.787  & 0.736  & ~ & 0.796  & 0.534  & ~ & 0.695   \\ 
           \cellcolor{white} ~ & 8k & 0.630  & 0.780  & 0.751  & ~ & 0.583  & 0.791  & 0.454  & 0.446  & 0.798  & 0.715  & ~ & 0.687  & 0.497  & ~ & 0.649   \\ 
           \cellcolor{white} ~ & 16k & 0.601  & 0.748  & 0.680  & ~ & 0.440  & 0.566  & 0.547  & 0.372  & 0.667  & 0.506  & ~ & 0.739  & 0.367  & ~ & 0.569   \\ 
           \cellcolor{white} ~ & 32k & 0.501  & 0.812  & 0.609  & ~ & 0.603  & 0.481  & 0.483  & 0.377  & 0.702  & 0.475  & ~ & 0.438  & 0.285  & ~ & 0.485   \\ 
           \cellcolor{white} ~ & 64k & 0.518  & 0.766  & 0.568  & ~ & 0.484  & 0.519  & 0.472  & 0.435  & 0.670  & 0.524  & ~ & 0.617  & 0.211  & ~ & 0.491   \\ 
           \multirow{-6}{*}{Qwen2.5-14B-Inst.\textsuperscript{$\dagger$}} & 128k & 0.463  & 0.740  & 0.484  & ~ & 0.341  & 0.510  & 0.308  & 0.318  & 0.577  & 0.468  & ~ & 0.269  & 0.203  & ~ & 0.394   \\ \hline
           \cellcolor{white} ~ & 4k & 0.608  & 0.809  & 0.492  & ~ & 0.688  & 0.455  & 0.691  & 0.751  & 0.712  & 0.834  & ~ & 0.734  & 0.682  & ~ & 0.648   \\ 
            \cellcolor{white} ~ & 8k & 0.490  & 0.869  & 0.540  & ~ & 0.675  & 0.384  & 0.592  & 0.682  & 0.678  & 0.868  & ~ & 0.744  & 0.625  & ~ & 0.617   \\ 
            \cellcolor{white} ~ & 16k & 0.399  & 0.841  & 0.439  & ~ & 0.738  & 0.434  & 0.530  & 0.637  & 0.717  & 0.727  & ~ & 0.710  & 0.338  & ~ & 0.523   \\ 
            \cellcolor{white} ~ & 32k & 0.386  & 0.847  & 0.329  & ~ & 0.638  & 0.449  & 0.594  & 0.704  & 0.683  & 0.803  & ~ & 0.767  & 0.334  & ~ & 0.518   \\ 
            \cellcolor{white} ~ & 64k & 0.435  & 0.821  & 0.213  & ~ & 0.488  & 0.489  & 0.475  & 0.551  & 0.762  & 0.721  & ~ & 0.227  & 0.313  & ~ & 0.414   \\ 
            \multirow{-6}{*}{InternLM2.5-7b-chat-1m} & 128k & 0.361  & 0.781  & 0.256  & ~ & 0.429  & 0.416  & 0.377  & 0.459  & 0.727  & 0.630  & ~ & 0.534  & 0.275  & ~ & 0.419   \\ \hline
            \cellcolor{white} ~ & 4k & 0.545  & 0.850  & 0.543  & ~ & 0.692  & 0.615  & 0.693  & 0.660  & 0.808  & 0.745  & ~ & 0.637  & 0.723  & ~ & 0.656   \\ 
            \cellcolor{white} ~ & 8k & 0.484  & 0.831  & 0.569  & ~ & 0.717  & 0.623  & 0.493  & 0.611  & 0.783  & 0.744  & ~ & 0.679  & 0.604  & ~ & 0.623   \\ 
            \cellcolor{white} ~ & 16k & 0.516  & 0.815  & 0.621  & ~ & 0.717  & 0.556  & 0.454  & 0.589  & 0.672  & 0.618  & ~ & 0.573  & 0.506  & ~ & 0.578   \\ 
            \cellcolor{white} ~ & 32k & 0.555  & 0.866  & 0.335  & ~ & 0.713  & 0.530  & 0.563  & 0.612  & 0.722  & 0.710  & ~ & 0.112  & 0.278  & ~ & 0.444   \\ 
            \cellcolor{white} ~ & 64k & 0.483  & 0.787  & 0.321  & ~ & 0.575  & 0.554  & 0.245  & 0.372  & 0.500  & 0.693  & ~ & 0.510  & 0.274  & ~ & 0.445   \\ 
            \multirow{-6}{*}{Qwen2.5-7B-Inst.\textsuperscript{$\dagger$}} & 128k & 0.462  & 0.724  & 0.292  & ~ & 0.346  & 0.531  & 0.166  & 0.340  & 0.619  & 0.698  & ~ & 0.156  & 0.230  & ~ & 0.366   \\
           \Xhline{1pt} 
       \end{tabular}
    }
\end{table}
\begin{table}[H]
    \small

    \rowcolors{2}{white}{gray!10}
    \resizebox{\textwidth}{!}{
    \begin{tabular}{c|ccccp{1pt}ccccccp{1pt}ccp{1pt}c}
     \arrayrulecolor{black}
        \Xhline{1pt}
        \rowcolor[gray]{0.9} 
        & & \multicolumn{3}{c}{\textbf{OneDoc}} & & \multicolumn{6}{c}{\textbf{List}} & & \multicolumn{2}{c}{\textbf{MultiDoc}} & &\\
        \noalign{\vskip -2pt} \cmidrule(lr){3-5} \cmidrule(lr){7-12}  \cmidrule(lr){14-15}
         \rowcolor[gray]{0.9} \noalign{\vskip -3pt}
        \multirow{-2}{*}{\textbf{Model}}& \multirow{-2}{*}{\textbf{Length}} &\textbf{OR} & \textbf{OQ} & \textbf{OE} & & \textbf{LSI} & \textbf{LMI} & \textbf{LOI} & \textbf{LOE} & \textbf{LBI} & \textbf{LBE} & & \textbf{MB} & \textbf{MF} & &\multirow{-2}{*}{\textbf{Overall}}\\  \hline
        \cellcolor{white} ~ & 4k & 0.713  & 0.609  & 0.532  & ~ & 0.721  & 0.680  & 0.437  & 0.607  & 0.739  & 0.540  & ~ & 0.569  & 0.436  & ~ & 0.580   \\ 
        \cellcolor{white} ~ & 8k & 0.547  & 0.734  & 0.353  & ~ & 0.754  & 0.668  & 0.392  & 0.545  & 0.639  & 0.599  & ~ & 0.630  & 0.372  & ~ & 0.527   \\ 
        \cellcolor{white} ~ & 16k & 0.624  & 0.679  & 0.517  & ~ & 0.783  & 0.536  & 0.467  & 0.613  & 0.731  & 0.731  & ~ & 0.638  & 0.292  & ~ & 0.548   \\ 
        \cellcolor{white} ~ & 32k & 0.426  & 0.648  & 0.167  & ~ & 0.642  & 0.530  & 0.399  & 0.557  & 0.646  & 0.655  & ~ & 0.623  & 0.340  & ~ & 0.457   \\ 
        \cellcolor{white} ~ & 64k & 0.464  & 0.543  & 0.552  & ~ & 0.725  & 0.428  & 0.403  & 0.530  & 0.703  & 0.688  & ~ & 0.356  & 0.219  & ~ & 0.446   \\ 
        \multirow{-6}{*}{Llama-3.1-8B-Inst.} & 128k & 0.449  & 0.874  & 0.355  & ~ & 0.604  & 0.366  & 0.287  & 0.373  & 0.548  & 0.612  & ~ & 0.317  & 0.191  & ~ & 0.388   \\ \hline
        \cellcolor{white} ~ & 4k & 0.578  & 0.837  & 0.388  & ~ & 0.738  & 0.692  & 0.425  & 0.633  & 0.702  & 0.749  & ~ & 0.743  & 0.441  & ~ & 0.589   \\ 
        \cellcolor{white} ~ & 8k & 0.450  & 0.877  & 0.262  & ~ & 0.788  & 0.616  & 0.415  & 0.511  & 0.671  & 0.747  & ~ & 0.742  & 0.431  & ~ & 0.540   \\ 
        \cellcolor{white} ~ & 16k & 0.498  & 0.746  & 0.322  & ~ & 0.825  & 0.499  & 0.428  & 0.463  & 0.623  & 0.613  & ~ & 0.795  & 0.299  & ~ & 0.509   \\ 
        \cellcolor{white} ~ & 32k & 0.503  & 0.751  & 0.212  & ~ & 0.717  & 0.468  & 0.375  & 0.555  & 0.670  & 0.661  & ~ & 0.784  & 0.252  & ~ & 0.482   \\ 
        \cellcolor{white} ~ & 64k & 0.423  & 0.830  & 0.166  & ~ & 0.688  & 0.484  & 0.314  & 0.467  & 0.646  & 0.604  & ~ & 0.713  & 0.196  & ~ & 0.437   \\ 
        \multirow{-6}{*}{GLM-4-9b-chat-1m} & 128k & 0.452  & 0.837  & 0.254  & ~ & 0.475  & 0.443  & 0.272  & 0.458  & 0.574  & 0.626  & ~ & 0.354  & 0.183  & ~ & 0.383   \\ \hline
        \cellcolor{white} ~ & 4k & 0.301  & 0.719  & 0.489  & ~ & 0.708  & 0.602  & 0.467  & 0.348  & 0.503  & 0.401  & ~ & 0.783  & 0.455  & ~ & 0.519   \\ 
        \cellcolor{white} ~ & 8k & 0.260  & 0.673  & 0.472  & ~ & 0.548  & 0.527  & 0.360  & 0.459  & 0.516  & 0.197  & ~ & 0.730  & 0.383  & ~ & 0.462   \\ 
        \cellcolor{white} ~ & 16k & 0.248  & 0.613  & 0.476  & ~ & 0.525  & 0.356  & 0.409  & 0.423  & 0.545  & 0.389  & ~ & 0.801  & 0.660  & ~ & 0.518   \\ 
        \cellcolor{white} ~ & 32k & 0.202  & 0.397  & 0.036  & ~ & 0.131  & 0.198  & 0.207  & 0.224  & 0.252  & 0.407  & ~ & 0.851  & 0.366  & ~ & 0.324   \\ 
        \cellcolor{white} ~ & 64k & 0.303  & 0.453  & 0.036  & ~ & 0.100  & 0.172  & 0.180  & 0.227  & 0.285  & 0.276  & ~ & 0.829  & 0.357  & ~ & 0.326   \\ 
        \multirow{-6}{*}{Qwen2.5-14B\textsuperscript{$\dagger$}} & 128k & 0.322  & 0.447  & 0.036  & ~ & 0.020  & 0.104  & 0.065  & 0.060  & 0.054  & 0.071  & ~ & 0.189  & 0.163  & ~ & 0.156   \\ \hline
        \cellcolor{white} ~ & 4k & 0.373  & 0.644  & 0.036  & ~ & 0.654  & 0.331  & 0.478  & 0.631  & 0.722  & 0.615  & ~ & 0.252  & 0.519  & ~ & 0.403   \\ 
        \cellcolor{white} ~ & 8k & 0.399  & 0.718  & 0.036  & ~ & 0.692  & 0.318  & 0.181  & 0.528  & 0.650  & 0.560  & ~ & 0.069  & 0.530  & ~ & 0.364   \\ 
        \cellcolor{white} ~ & 16k & 0.426  & 0.721  & 0.261  & ~ & 0.638  & 0.264  & 0.141  & 0.595  & 0.432  & 0.571  & ~ & 0.129  & 0.548  & ~ & 0.395   \\ 
        \cellcolor{white} ~ & 32k & 0.442  & 0.740  & 0.036  & ~ & 0.671  & 0.361  & 0.562  & 0.643  & 0.717  & 0.597  & ~ & 0.142  & 0.520  & ~ & 0.409   \\ 
        \cellcolor{white} ~ & 64k & 0.405  & 0.790  & 0.044  & ~ & 0.629  & 0.231  & 0.292  & 0.656  & 0.591  & 0.639  & ~ & 0.087  & 0.502  & ~ & 0.366   \\ 
        \multirow{-6}{*}{LWM-Text-Chat-1M} & 128k & 0.435  & 0.770  & 0.036  & ~ & 0.517  & 0.243  & 0.198  & 0.576  & 0.426  & 0.558  & ~ & 0.087  & 0.503  & ~ & 0.346   \\ \hline
        \cellcolor{white} ~ & 4k & 0.473  & 0.251  & 0.014  & ~ & 0.733  & 0.361  & 0.681  & 0.747  & 0.773  & 0.722  & ~ & 0.788  & 0.635  & ~ & 0.521   \\ 
        \cellcolor{white} ~ & 8k & 0.270  & 0.273  & 0.069  & ~ & 0.733  & 0.231  & 0.604  & 0.611  & 0.679  & 0.613  & ~ & 0.777  & 0.507  & ~ & 0.441   \\ 
        \cellcolor{white} ~ & 16k & 0.415  & 0.287  & 0.017  & ~ & 0.753  & 0.151  & 0.447  & 0.455  & 0.660  & 0.607  & ~ & 0.705  & 0.404  & ~ & 0.403   \\ 
        \cellcolor{white} ~ & 32k & 0.357  & 0.267  & 0.057  & ~ & 0.529  & 0.142  & 0.344  & 0.562  & 0.636  & 0.652  & ~ & 0.166  & 0.388  & ~ & 0.311   \\ 
        \cellcolor{white} ~ & 64k & 0.276  & 0.340  & 0.056  & ~ & 0.513  & 0.114  & 0.413  & 0.532  & 0.588  & 0.600  & ~ & 0.188  & 0.381  & ~ & 0.298   \\ 
        \multirow{-6}{*}{Llama-3.1-8B} & 128k & 0.290  & 0.307  & 0.030  & ~ & 0.340  & 0.243  & 0.042  & 0.414  & 0.414  & 0.541  & ~ & 0.200  & 0.415  & ~ & 0.278   \\ \hline
        \cellcolor{white} ~ & 4k & 0.192  & 0.624  & 0.450  & ~ & 0.033  & 0.100  & 0.000  & 0.026  & 0.000  & 0.000  & ~ & 0.363  & 0.388  & ~ & 0.264   \\ 
        \cellcolor{white} ~ & 8k & 0.214  & 0.665  & 0.353  & ~ & 0.084  & 0.103  & 0.022  & 0.103  & 0.020  & 0.073  & ~ & 0.633  & 0.233  & ~ & 0.274   \\ 
        \cellcolor{white} ~ & 16k & 0.245  & 0.301  & 0.438  & ~ & 0.314  & 0.113  & 0.226  & 0.311  & 0.393  & 0.299  & ~ & 0.000  & 0.183  & ~ & 0.229   \\ 
        \cellcolor{white} ~ & 32k & 0.323  & 0.470  & 0.036  & ~ & 0.000  & 0.110  & 0.013  & 0.013  & 0.016  & 0.007  & ~ & 0.371  & 0.219  & ~ & 0.184   \\ 
        \cellcolor{white} ~ & 64k & 0.298  & 0.565  & 0.087  & ~ & 0.125  & 0.136  & 0.118  & 0.125  & 0.229  & 0.334  & ~ & 0.075  & 0.209  & ~ & 0.193   \\ 
        \multirow{-6}{*}{Qwen2.5-7B\textsuperscript{$\dagger$}} & 128k & 0.336  & 0.320  & 0.036  & ~ & 0.083  & 0.116  & 0.031  & 0.035  & 0.055  & 0.179  & ~ & 0.021  & 0.164  & ~ & 0.134   \\ \hline
        \cellcolor{white} ~ & 4k & 0.334  & 0.293  & 0.036  & ~ & 0.259  & 0.126  & 0.233  & 0.311  & 0.565  & 0.428  & ~ & 0.151  & 0.230  & ~ & 0.230   \\ 
        \cellcolor{white} ~ & 8k & 0.315  & 0.187  & 0.069  & ~ & 0.240  & 0.109  & 0.081  & 0.359  & 0.466  & 0.471  & ~ & 0.151  & 0.241  & ~ & 0.220   \\ 
        \cellcolor{white} ~ & 16k & 0.279  & 0.240  & 0.050  & ~ & 0.116  & 0.125  & 0.081  & 0.335  & 0.307  & 0.450  & ~ & 0.113  & 0.194  & ~ & 0.186   \\ 
        \cellcolor{white} ~ & 32k & 0.325  & 0.261  & 0.036  & ~ & 0.148  & 0.139  & 0.149  & 0.280  & 0.482  & 0.414  & ~ & 0.111  & 0.266  & ~ & 0.216   \\ 
        \cellcolor{white} ~ & 64k & 0.294  & 0.267  & 0.059  & ~ & 0.178  & 0.113  & 0.118  & 0.304  & 0.512  & 0.519  & ~ & 0.128  & 0.190  & ~ & 0.208   \\ 
        \multirow{-6}{*}{LWM-Text-1M} & 128k & 0.293  & 0.267  & 0.044  & ~ & 0.042  & 0.205  & 0.000  & 0.250  & 0.189  & 0.429  & ~ & 0.016  & 0.197  & ~ & 0.167   \\ \hline
        \Xhline{1pt} 
    \end{tabular}
    }
    \caption{ARS scores in different input lengths $l$. $\dagger$ indicates that the context $X$ on the longest interval is right-truncated.}
    \label{tab:IFS_Full_Res}

\end{table}
\subsection{Instruction Following Performance in Six Capabilities}
\begin{table}[H]
    \centering
    \small
    \rowcolors{1}{white}{gray!10}
    \resizebox{12cm}{!}{
    \begin{tabular}{lcccccc}
        \arrayrulecolor{black}
       \Xhline{1pt}
       \rowcolor[gray]{0.9} 
        Models &\textbf{Format} & \textbf{Logit} & \textbf{Num} & \textbf{Ori} & \textbf{Recog} & \textbf{Spat}\\  \hline
        GPT-4o  & 0.712  & 0.582  & 0.842  & 0.836  & 0.756  & 0.806   \\
        GPT-4 & 0.789  & 0.509  & 0.765  & 0.921  & 0.615  & 0.729   \\
        Qwen2.5-72B-Inst.\textsuperscript{$\dagger$} & 0.727  & 0.468  & 0.703  & 0.840  & 0.640  & 0.718   \\
        Llama-3.1-70B-Inst. & 0.766  & 0.509  & 0.684  & 0.747  & 0.626  & 0.699   \\
        Qwen2.5-32B-Inst.\textsuperscript{$\dagger$} & 0.760  & 0.400  & 0.671  & 0.752  & 0.492  & 0.647   \\
        C4AI-cmd-r-08-2024 (32B)\textsuperscript{$\dagger$} & 0.721  & 0.480  & 0.678  & 0.833  & 0.453  & 0.508   \\
        C4AI-cmd-r-v01 (35B)\textsuperscript{$\dagger$} & 0.767  & 0.400  & 0.596  & 0.811  & 0.379  & 0.470   \\
        Qwen2.5-72B\textsuperscript{$\dagger$} & 0.615  & 0.344  & 0.526  & 0.770  & 0.408  & 0.524   \\
        Llama-3.1-70B & 0.492  & 0.253  & 0.437  & 0.594  & 0.290  & 0.410   \\
        Qwen2.5-32B\textsuperscript{$\dagger$} & 0.432  & 0.342  & 0.483  & 0.537  & 0.303  & 0.332   \\
        Qwen2.5-14B-Inst.\textsuperscript{$\dagger$} & 0.659  & 0.400  & 0.612  & 0.591  & 0.410  & 0.523   \\
        InternLM2.5-7b-chat-1m & 0.727  & 0.401  & 0.576  & 0.644  & 0.275  & 0.438   \\
        Qwen2.5-7B-Inst.\textsuperscript{$\dagger$} & 0.707  & 0.359  & 0.612  & 0.595  & 0.317  & 0.420   \\
        Llama-3.1-8B-Inst. & 0.465  & 0.349  & 0.539  & 0.627  & 0.393  & 0.553   \\
        GLM-4-9b-chat-1m & 0.672  & 0.363  & 0.604  & 0.548  & 0.329  & 0.364   \\
        Qwen2.5-14B\textsuperscript{$\dagger$} & 0.420  & 0.318  & 0.401  & 0.574  & 0.307  & 0.318   \\
        LWM-Text-Chat-1M & 0.505  & 0.265  & 0.370  & 0.578  & 0.186  & 0.312   \\
        Llama-3.1-8B & 0.586  & 0.328  & 0.331  & 0.521  & 0.088  & 0.303   \\
        Qwen2.5-7B\textsuperscript{$\dagger$} & 0.274  & 0.254  & 0.229  & 0.274  & 0.117  & 0.206   \\
        LWM-Text-1M & 0.293  & 0.105  & 0.103  & 0.374  & 0.044  & 0.208   \\
       \Xhline{1pt} 
       \end{tabular}
    }
    \caption{Instruction following performance in six core capabilities. $\dagger$ indicates that the context $X$ on the longest interval is right-truncated.}
\end{table}

\clearpage
\begin{table}[H]
    \renewcommand{\arraystretch}{1.3}
    \rowcolors{2}{white}{gray!10}
    \resizebox{\textwidth}{!}{
    \begin{tabular}{p{19ex}<{\centering}|ccccccc||p{22ex}<{\centering}|ccccccc}
        \arrayrulecolor{black}
       \Xhline{1pt}
       \rowcolor[gray]{0.9} 
        \textbf{Models} & \textbf{Length} &\textbf{Format} & \textbf{Logit} & \textbf{Num} & \textbf{Ori} & \textbf{Recog} & \textbf{Spat} &\textbf{Models} & \textbf{Length} &\textbf{Format} & \textbf{Logit} & \textbf{Num} & \textbf{Ori} & \textbf{Recog} & \textbf{Spat}\\  \hline
            \cellcolor{white} & 4k & 0.683 & 0.519 & 0.833 & 0.851 & 0.896 & 0.853 & \cellcolor{white} & 4k & 0.786 & 0.653 & 0.827 & 0.999 & 0.900 & 0.889\\
            \cellcolor{white} & 8k & 0.712 & 0.630 & 0.859 & 0.852 & 0.909 & 0.826 & \cellcolor{white} & 8k & 0.809 & 0.701 & 0.850 & 0.997 & 0.840 & 0.860\\
            \cellcolor{white} & 16k & 0.715 & 0.619 & 0.873 & 0.906 & 0.836 & 0.846 & \cellcolor{white} & 16k & 0.771 & 0.545 & 0.851 & 0.983 & 0.792 & 0.815\\
            \cellcolor{white} & 32k & 0.732 & 0.633 & 0.887 & 0.866 & 0.741 & 0.834 & \cellcolor{white} & 32k & 0.813 & 0.425 & 0.744 & 0.885 & 0.526 & 0.723\\
            \cellcolor{white} & 64k & 0.737 & 0.592 & 0.783 & 0.792 & 0.630 & 0.773 & \cellcolor{white} & 64k & 0.839 & 0.373 & 0.666 & 0.896 & 0.377 & 0.564\\
            \multirow{-6}{*}{GPT-4o} & 128k & 0.693 & 0.502 & 0.816 & 0.751 & 0.520 & 0.706 & \multirow{-6}{*}{GPT-4} & 128k & 0.719 & 0.358 & 0.654 & 0.768 & 0.257 & 0.524\\\hline 
            \cellcolor{white} & 4k & 0.786 & 0.564 & 0.817 & 0.994 & 0.883 & 0.837 & \cellcolor{white} & 4k & 0.793 & 0.661 & 0.762 & 0.852 & 0.831 & 0.840\\
            \cellcolor{white} & 8k & 0.798 & 0.545 & 0.761 & 0.953 & 0.806 & 0.818 & \cellcolor{white} & 8k & 0.791 & 0.680 & 0.857 & 0.889 & 0.899 & 0.891\\
            \cellcolor{white} & 16k & 0.771 & 0.424 & 0.708 & 0.962 & 0.682 & 0.720 & \cellcolor{white} & 16k & 0.808 & 0.650 & 0.790 & 0.882 & 0.761 & 0.848\\
            \cellcolor{white} & 32k & 0.741 & 0.437 & 0.724 & 0.892 & 0.584 & 0.734 & \cellcolor{white} & 32k & 0.798 & 0.365 & 0.632 & 0.829 & 0.610 & 0.780\\
            \cellcolor{white} & 64k & 0.667 & 0.562 & 0.653 & 0.741 & 0.517 & 0.618 & \cellcolor{white} & 64k & 0.784 & 0.474 & 0.592 & 0.745 & 0.463 & 0.627\\
            \multirow{-6}{*}{Qwen2.5-72B-Inst.\textsuperscript{$\dagger$}} & 128k & 0.599 & 0.276 & 0.555 & 0.503 & 0.369 & 0.581 & \multirow{-6}{*}{Llama-3.1-70B-Inst.} & 128k & 0.625 & 0.223 & 0.470 & 0.283 & 0.193 & 0.206\\\hline 
            \cellcolor{white} & 4k & 0.893 & 0.548 & 0.793 & 0.977 & 0.710 & 0.715 & \cellcolor{white} & 4k & 0.844 & 0.659 & 0.868 & 0.959 & 0.707 & 0.727\\
            \cellcolor{white} & 8k & 0.864 & 0.402 & 0.740 & 0.898 & 0.643 & 0.704 & \cellcolor{white} & 8k & 0.823 & 0.661 & 0.882 & 0.879 & 0.622 & 0.655\\
            \cellcolor{white} & 16k & 0.793 & 0.409 & 0.693 & 0.842 & 0.507 & 0.652 & \cellcolor{white} & 16k & 0.729 & 0.481 & 0.666 & 0.959 & 0.522 & 0.476\\
            \cellcolor{white} & 32k & 0.643 & 0.375 & 0.599 & 0.628 & 0.425 & 0.642 & \cellcolor{white} & 32k & 0.729 & 0.458 & 0.617 & 0.863 & 0.400 & 0.491\\
            \cellcolor{white} & 64k & 0.706 & 0.406 & 0.615 & 0.673 & 0.399 & 0.648 & \cellcolor{white} & 64k & 0.631 & 0.361 & 0.533 & 0.775 & 0.291 & 0.379\\
            \multirow{-6}{*}{Qwen2.5-32B-Inst.\textsuperscript{$\dagger$}} & 128k & 0.662 & 0.261 & 0.583 & 0.495 & 0.268 & 0.520 & \multirow{-6}{*}{C4AI-cmd-r-08-2024\textsuperscript{$\dagger$}} & 128k & 0.573 & 0.260 & 0.503 & 0.562 & 0.175 & 0.321\\\hline 
            \cellcolor{white} & 4k & 0.759 & 0.625 & 0.767 & 0.944 & 0.606 & 0.555 & \cellcolor{white} & 4k & 0.692 & 0.465 & 0.699 & 0.964 & 0.685 & 0.727\\
            \cellcolor{white} & 8k & 0.754 & 0.505 & 0.730 & 0.891 & 0.515 & 0.501 & \cellcolor{white} & 8k & 0.629 & 0.322 & 0.509 & 0.928 & 0.643 & 0.652\\
            \cellcolor{white} & 16k & 0.742 & 0.376 & 0.553 & 0.825 & 0.358 & 0.462 & \cellcolor{white} & 16k & 0.577 & 0.394 & 0.561 & 0.813 & 0.421 & 0.596\\
            \cellcolor{white} & 32k & 0.797 & 0.353 & 0.562 & 0.856 & 0.331 & 0.469 & \cellcolor{white} & 32k & 0.613 & 0.323 & 0.485 & 0.847 & 0.318 & 0.533\\
            \cellcolor{white} & 64k & 0.779 & 0.302 & 0.477 & 0.742 & 0.273 & 0.464 & \cellcolor{white} & 64k & 0.630 & 0.318 & 0.477 & 0.578 & 0.219 & 0.308\\
            \multirow{-6}{*}{C4AI-cmd-r-v01\textsuperscript{$\dagger$}} & 128k & 0.772 & 0.238 & 0.487 & 0.607 & 0.193 & 0.367 & \multirow{-6}{*}{Qwen2.5-72B\textsuperscript{$\dagger$}} & 128k & 0.548 & 0.245 & 0.424 & 0.489 & 0.159 & 0.330\\\hline 
            \cellcolor{white} & 4k & 0.457 & 0.353 & 0.558 & 0.589 & 0.430 & 0.502 & \cellcolor{white} & 4k & 0.498 & 0.477 & 0.679 & 0.771 & 0.550 & 0.616\\
            \cellcolor{white} & 8k & 0.447 & 0.258 & 0.452 & 0.618 & 0.342 & 0.363 & \cellcolor{white} & 8k & 0.560 & 0.459 & 0.713 & 0.839 & 0.448 & 0.615\\
            \cellcolor{white} & 16k & 0.549 & 0.250 & 0.512 & 0.644 & 0.340 & 0.405 & \cellcolor{white} & 16k & 0.457 & 0.440 & 0.491 & 0.508 & 0.305 & 0.385\\
            \cellcolor{white} & 32k & 0.538 & 0.268 & 0.452 & 0.643 & 0.308 & 0.512 & \cellcolor{white} & 32k & 0.432 & 0.345 & 0.367 & 0.456 & 0.241 & 0.124\\
            \cellcolor{white} & 64k & 0.489 & 0.203 & 0.385 & 0.570 & 0.251 & 0.394 & \cellcolor{white} & 64k & 0.419 & 0.243 & 0.448 & 0.476 & 0.190 & 0.138\\
            \multirow{-6}{*}{Llama-3.1-70B} & 128k & 0.473 & 0.186 & 0.263 & 0.499 & 0.070 & 0.281 & \multirow{-6}{*}{Qwen2.5-32B\textsuperscript{$\dagger$}} & 128k & 0.223 & 0.088 & 0.201 & 0.172 & 0.082 & 0.115\\\hline 
            \cellcolor{white} & 4k & 0.733 & 0.628 & 0.767 & 0.729 & 0.633 & 0.622 & \cellcolor{white} & 4k & 0.815 & 0.583 & 0.746 & 0.811 & 0.417 & 0.542\\
            \cellcolor{white} & 8k & 0.717 & 0.527 & 0.677 & 0.689 & 0.554 & 0.665 & \cellcolor{white} & 8k & 0.813 & 0.550 & 0.710 & 0.753 & 0.373 & 0.520\\
            \cellcolor{white} & 16k & 0.676 & 0.415 & 0.595 & 0.647 & 0.462 & 0.527 & \cellcolor{white} & 16k & 0.771 & 0.350 & 0.553 & 0.618 & 0.290 & 0.424\\
            \cellcolor{white} & 32k & 0.649 & 0.293 & 0.554 & 0.501 & 0.339 & 0.458 & \cellcolor{white} & 32k & 0.734 & 0.355 & 0.554 & 0.648 & 0.264 & 0.439\\
            \cellcolor{white} & 64k & 0.653 & 0.296 & 0.565 & 0.540 & 0.291 & 0.492 & \cellcolor{white} & 64k & 0.633 & 0.277 & 0.414 & 0.505 & 0.116 & 0.353\\
            \multirow{-6}{*}{Qwen2.5-14B-Inst.\textsuperscript{$\dagger$}} & 128k & 0.527 & 0.240 & 0.513 & 0.404 & 0.180 & 0.377 & \multirow{-6}{*}{InternLM2.5-7b-chat-1m} & 128k & 0.594 & 0.289 & 0.480 & 0.531 & 0.190 & 0.352\\\hline 
            \cellcolor{white} & 4k & 0.841 & 0.487 & 0.692 & 0.810 & 0.470 & 0.536 & \cellcolor{white} & 4k & 0.413 & 0.457 & 0.605 & 0.737 & 0.590 & 0.685\\
            \cellcolor{white} & 8k & 0.813 & 0.465 & 0.752 & 0.759 & 0.399 & 0.503 & \cellcolor{white} & 8k & 0.460 & 0.465 & 0.628 & 0.623 & 0.503 & 0.545\\
            \cellcolor{white} & 16k & 0.749 & 0.381 & 0.628 & 0.728 & 0.401 & 0.456 & \cellcolor{white} & 16k & 0.519 & 0.373 & 0.563 & 0.680 & 0.471 & 0.623\\
            \cellcolor{white} & 32k & 0.644 & 0.307 & 0.506 & 0.426 & 0.247 & 0.387 & \cellcolor{white} & 32k & 0.476 & 0.332 & 0.549 & 0.686 & 0.274 & 0.423\\
            \cellcolor{white} & 64k & 0.652 & 0.270 & 0.600 & 0.486 & 0.245 & 0.321 & \cellcolor{white} & 64k & 0.436 & 0.252 & 0.460 & 0.577 & 0.291 & 0.623\\
            \multirow{-6}{*}{Qwen2.5-7B-Inst.\textsuperscript{$\dagger$}} & 128k & 0.545 & 0.242 & 0.496 & 0.361 & 0.141 & 0.317 & \multirow{-6}{*}{Llama-3.1-8B-Inst.} & 128k & 0.485 & 0.217 & 0.429 & 0.456 & 0.230 & 0.418\\\hline 
            \cellcolor{white} & 4k & 0.726 & 0.482 & 0.727 & 0.629 & 0.461 & 0.488 & \cellcolor{white} & 4k & 0.413 & 0.394 & 0.533 & 0.739 & 0.553 & 0.500\\
            \cellcolor{white} & 8k & 0.716 & 0.458 & 0.682 & 0.579 & 0.423 & 0.374 & \cellcolor{white} & 8k & 0.394 & 0.390 & 0.528 & 0.667 & 0.432 & 0.476\\
            \cellcolor{white} & 16k & 0.713 & 0.358 & 0.624 & 0.605 & 0.341 & 0.332 & \cellcolor{white} & 16k & 0.491 & 0.363 & 0.459 & 0.886 & 0.395 & 0.504\\
            \cellcolor{white} & 32k & 0.672 & 0.323 & 0.591 & 0.584 & 0.286 & 0.350 & \cellcolor{white} & 32k & 0.367 & 0.322 & 0.405 & 0.561 & 0.206 & 0.192\\
            \cellcolor{white} & 64k & 0.626 & 0.312 & 0.522 & 0.498 & 0.261 & 0.321 & \cellcolor{white} & 64k & 0.525 & 0.275 & 0.335 & 0.458 & 0.182 & 0.158\\
            \multirow{-6}{*}{GLM-4-9b-chat-1m} & 128k & 0.578 & 0.246 & 0.477 & 0.391 & 0.199 & 0.319 & \multirow{-6}{*}{Qwen2.5-14B\textsuperscript{$\dagger$}} & 128k & 0.332 & 0.166 & 0.143 & 0.132 & 0.076 & 0.079\\\hline 
            \cellcolor{white} & 4k & 0.640 & 0.313 & 0.370 & 0.534 & 0.107 & 0.334 & \cellcolor{white} & 4k & 0.570 & 0.425 & 0.640 & 0.784 & 0.379 & 0.408\\
            \cellcolor{white} & 8k & 0.557 & 0.329 & 0.345 & 0.506 & 0.081 & 0.251 & \cellcolor{white} & 8k & 0.555 & 0.331 & 0.497 & 0.688 & 0.299 & 0.322\\
            \cellcolor{white} & 16k & 0.570 & 0.365 & 0.351 & 0.548 & 0.087 & 0.361 & \cellcolor{white} & 16k & 0.495 & 0.325 & 0.451 & 0.645 & 0.253 & 0.278\\
            \cellcolor{white} & 32k & 0.625 & 0.333 & 0.386 & 0.543 & 0.103 & 0.331 & \cellcolor{white} & 32k & 0.477 & 0.178 & 0.250 & 0.464 & 0.065 & 0.317\\
            \cellcolor{white} & 64k & 0.573 & 0.318 & 0.258 & 0.517 & 0.083 & 0.293 & \cellcolor{white} & 64k & 0.467 & 0.167 & 0.176 & 0.461 & 0.069 & 0.311\\
            \multirow{-6}{*}{LWM-Text-Chat-1M} & 128k & 0.554 & 0.311 & 0.275 & 0.480 & 0.069 & 0.245 & \multirow{-6}{*}{Llama-3.1-8B} & 128k & 0.463 & 0.161 & 0.205 & 0.423 & 0.051 & 0.236\\\hline 
            \cellcolor{white} & 4k & 0.244 & 0.389 & 0.351 & 0.290 & 0.236 & 0.254 & \cellcolor{white} & 4k & 0.317 & 0.145 & 0.138 & 0.424 & 0.045 & 0.233\\
            \cellcolor{white} & 8k & 0.307 & 0.384 & 0.389 & 0.350 & 0.200 & 0.236 & \cellcolor{white} & 8k & 0.288 & 0.034 & 0.034 & 0.469 & 0.061 & 0.249\\
            \cellcolor{white} & 16k & 0.183 & 0.161 & 0.129 & 0.428 & 0.070 & 0.404 & \cellcolor{white} & 16k & 0.280 & 0.099 & 0.095 & 0.325 & 0.042 & 0.176\\
            \cellcolor{white} & 32k & 0.334 & 0.247 & 0.250 & 0.185 & 0.114 & 0.066 & \cellcolor{white} & 32k & 0.300 & 0.135 & 0.129 & 0.417 & 0.041 & 0.196\\
            \cellcolor{white} & 64k & 0.314 & 0.209 & 0.146 & 0.233 & 0.047 & 0.182 & \cellcolor{white} & 64k & 0.310 & 0.102 & 0.077 & 0.362 & 0.034 & 0.238\\
            \multirow{-6}{*}{Qwen2.5-7B\textsuperscript{$\dagger$}} & 128k & 0.260 & 0.132 & 0.109 & 0.161 & 0.038 & 0.095 & \multirow{-6}{*}{LWM-Text-1M} & 128k & 0.264 & 0.116 & 0.147 & 0.250 & 0.044 & 0.155\\\hline 
           \Xhline{1pt} 
       \end{tabular}
       }
       \caption{Instruction following performance in different input lengths $l$. $\dagger$ indicates that the context $X$ on the longest interval is right-truncated.}
       \label{tab:IFP-Len}
\end{table}
\clearpage
\subsection{Instruction Following Stability}
\begin{table}[H]
    \centering
    %
    \rowcolors{2}{white}{gray!10}
    \resizebox{12cm}{!}{
    \begin{tabular}{l|cccccc}
        \arrayrulecolor{black}
       \Xhline{1pt}
       \rowcolor[gray]{0.9} 
        \textbf{Models} & \textbf{4k} & \textbf{8k} & \textbf{16k} & \textbf{32k} & \textbf{64k} & \textbf{128k} \\  \hline
        GPT-4o & 0.102  & 0.124  & 0.100  & 0.124  & 0.140  & 0.155   \\
        GPT-4 & 0.073  & 0.075  & 0.086  & 0.116  & 0.121  & 0.188   \\ 
        Qwen2.5-72B-Inst.\textsuperscript{$\dagger$} & 0.100  & 0.103  & 0.092  & 0.094  & 0.120  & 0.136   \\ 
        Llama-3.1-70B-Inst. & 0.127  & 0.124  & 0.118  & 0.154  & 0.150  & 0.261   \\ 
        Qwen2.5-32B-Inst.\textsuperscript{$\dagger$} & 0.109  & 0.137  & 0.152  & 0.164  & 0.152  & 0.169   \\ 
        C4AI-cmd-r-08-2024\textsuperscript{$\dagger$} & 0.121  & 0.133  & 0.177  & 0.180  & 0.229  & 0.226   \\ 
        C4AI-cmd-r-v01\textsuperscript{$\dagger$} & 0.171  & 0.191  & 0.233  & 0.185  & 0.191  & 0.189   \\ 
        Qwen2.5-72B\textsuperscript{$\dagger$} & 0.193  & 0.227  & 0.207  & 0.214  & 0.319  & 0.243   \\ 
        Llama-3.1-70B & 0.485  & 0.510  & 0.301  & 0.311  & 0.301  & 0.235   \\ 
        Qwen2.5-32B\textsuperscript{$\dagger$} & 0.218  & 0.224  & 0.749  & 1.250  & 0.426  & 0.521   \\ 
        Qwen2.5-14B-Inst.\textsuperscript{$\dagger$} & 0.150  & 0.184  & 0.181  & 0.207  & 0.187  & 0.196   \\ 
        InternLM2.5-7b-chat-1m & 0.145  & 0.157  & 0.152  & 0.173  & 0.202  & 0.265   \\ 
        Qwen2.5-7B-Inst.\textsuperscript{$\dagger$} & 0.153  & 0.146  & 0.181  & 0.217  & 0.259  & 0.296   \\ 
        Llama-3.1-8B-Inst. & 0.221  & 0.244  & 0.203  & 0.271  & 0.228  & 0.225   \\ 
        GLM-4-9b-chat-1m & 0.165  & 0.163  & 0.195  & 0.214  & 0.254  & 0.229   \\ 
        Qwen2.5-14B\textsuperscript{$\dagger$} & 0.249  & 0.271  & 0.230  & 0.521  & 0.515  & 0.806   \\ 
        LWM-Text-Chat-1M & 0.164  & 0.341  & 0.334  & 0.196  & 0.373  & 0.379   \\ 
        Llama-3.1-8B & 0.356  & 0.349  & 0.276  & 0.330  & 0.322  & 0.406   \\ 
        Qwen2.5-7B\textsuperscript{$\dagger$} & 0.740  & 0.877  & 0.319  & 1.418  & 0.568  & 0.866   \\ 
        LWM-Text-1M & 0.200  & 0.366  & 0.469  & 0.333  & 0.349  & 0.608   \\
           
           \Xhline{1pt} 
       \end{tabular}
       }
       \caption{Instruction following stability (Expression) in different input lengths $l$. $\dagger$ indicates that the context $X$ on the longest interval is right-truncated.}
       \label{tab:IFS-Expression-Len}
\end{table}

\begin{table}[H]
    \centering
    %
    \rowcolors{2}{white}{gray!10}
    \resizebox{12cm}{!}{
    \begin{tabular}{l|cccccc}
        \arrayrulecolor{black}
       \Xhline{1pt}
       \rowcolor[gray]{0.9} 
        \textbf{Models} & \textbf{4k} & \textbf{8k} & \textbf{16k} & \textbf{32k} & \textbf{64k} & \textbf{128k} \\  \hline
        GPT-4o & 0.093  & 0.085  & 0.101  & 0.105  & 0.117  & 0.164   \\
        GPT-4 & 0.089  & 0.097  & 0.119  & 0.145  & 0.183  & 0.187   \\ 
        Qwen2.5-72B-Inst.\textsuperscript{$\dagger$} & 0.074  & 0.108  & 0.132  & 0.141  & 0.179  & 0.142   \\ 
        Llama-3.1-70B-Inst. & 0.113  & 0.075  & 0.125  & 0.122  & 0.198  & 0.273   \\ 
        Qwen2.5-32B-Inst.\textsuperscript{$\dagger$} & 0.136  & 0.132  & 0.177  & 0.165  & 0.152  & 0.184   \\ 
        C4AI-cmd-r-08-2024\textsuperscript{$\dagger$} & 0.140  & 0.156  & 0.148  & 0.156  & 0.205  & 0.147   \\ 
        C4AI-cmd-r-v01\textsuperscript{$\dagger$} & 0.167  & 0.146  & 0.151  & 0.184  & 0.175  & 0.196   \\ 
        Qwen2.5-72B\textsuperscript{$\dagger$} & 0.143  & 0.120  & 0.168  & 0.223  & 0.271  & 0.211   \\ 
        Llama-3.1-70B & 0.405  & 0.500  & 0.272  & 0.183  & 0.172  & 0.220   \\ 
        Qwen2.5-32B\textsuperscript{$\dagger$} & 0.203  & 0.146  & 0.302  & 0.600  & 0.254  & 0.483   \\ 
        Qwen2.5-14B-Inst.\textsuperscript{$\dagger$} & 0.162  & 0.186  & 0.193  & 0.161  & 0.192  & 0.145   \\ 
        InternLM2.5-7b-chat-1m & 0.191  & 0.155  & 0.181  & 0.217  & 0.336  & 0.240   \\ 
        Qwen2.5-7B-Inst.\textsuperscript{$\dagger$} & 0.156  & 0.169  & 0.204  & 0.216  & 0.252  & 0.187   \\ 
        Llama-3.1-8B-Inst. & 0.149  & 0.167  & 0.183  & 0.320  & 0.212  & 0.249   \\ 
        GLM-4-9b-chat-1m & 0.205  & 0.195  & 0.187  & 0.227  & 0.230  & 0.239   \\ 
        Qwen2.5-14B\textsuperscript{$\dagger$} & 0.200  & 0.214  & 0.182  & 0.233  & 0.308  & 0.572   \\ 
        LWM-Text-Chat-1M & 0.158  & 0.159  & 0.256  & 0.195  & 0.234  & 0.206   \\ 
        Llama-3.1-8B & 0.384  & 0.233  & 0.268  & 0.211  & 0.235  & 0.427   \\ 
        Qwen2.5-7B\textsuperscript{$\dagger$} & 0.456  & 0.420  & 0.145  & 0.816  & 0.364  & 0.499   \\ 
        LWM-Text-1M & 0.158  & 0.379  & 0.331  & 0.281  & 0.311  & 0.533   \\ 
        \Xhline{1pt} 
       \end{tabular}
       }
       \caption{Instruction following stability (Variable) in different input lengths $l$. $\dagger$ indicates that the context $X$ on the longest interval is right-truncated.}
       \label{tab:IFS-Variable-Len}
\end{table}
\newpage
\section{Examples in \ours}
\label{sec:exampleinLIFBench}
\small
\begin{longtable}{p{3cm}p{11.5cm}}
    \hline
    \textbf{Task Name}           & \textbf{Test Example}  \\
    \hline
    \endfirsthead
    \hline
    \textbf{Task Name}           & \textbf{Test Example}  \\
    \hline
    \endhead
    
    Scenario:  \textit{List} \newline Task:  \textit{Single-ID} & \textbf{You're a searcher. You need to output the corresponding list elements based on the instructions and the list below. Please follow the instructions directly without anything else.} \newline List to be retrieved:\newline 1. 4f63efbe7f5111ef8b42581122bf941e\newline 2.Summarize the Rome Statute of the International Criminal Court.\newline  ...\newline 153. 4f7ea64c7f5111ef8b42581122bf941e\newline \textit{Instruction: Retrieve the entry at position \textcolor{red}{8th} in the list. Display it immediately.}\\
    \hline
    Scenario:  \textit{List} \newline Task:  \textit{Multi-ID} & \textbf{You're a searcher. You need to output the corresponding list elements based on the instructions and the list below. Please follow the instructions directly without anything else.} \newline List to be retrieved:\newline 1. 4f63efbe7f5111ef8b42581122bf941e\newline 2.Summarize the Rome Statute of the International Criminal Court.\newline  ...\newline 153. 4f7ea64c7f5111ef8b42581122bf941e\newline \textit{Instruction: Identify the items at the corresponding places in list \textcolor{red}{[1, 5, 7]} and deliver the result in JSON list form.}\\
    \hline
    Scenario:  \textit{List} \newline Task:  \textit{Offset-ID} & \textbf{You're a searcher. You need to output the corresponding list elements based on the instructions and the list below. Please follow the instructions directly without anything else.} \newline List to be retrieved:\newline 1. 4f63efbe7f5111ef8b42581122bf941e\newline 2.Summarize the Rome Statute of the International Criminal Court.\newline  ...\newline 153. 4f7ea64c7f5111ef8b42581122bf941e\newline \textit{Instruction: Please identify \textcolor{red}{the next} item to \textcolor{red}{8th} in the list described above.}\\
    \hline
    Scenario:  \textit{List} \newline Task:  \textit{Offset-Element} & \textbf{You're a searcher. You need to output the corresponding list elements based on the instructions and the list below. Please follow the instructions directly without anything else.} \newline List to be retrieved:\newline 1. 4f63efbe7f5111ef8b42581122bf941e\newline 2.Summarize the Rome Statute of the International Criminal Court.\newline  ...\newline 153. 4f7ea64c7f5111ef8b42581122bf941e\newline \textit{Instruction: "Considering the arrangement of the list, what is \textcolor{red}{the next} element after \textcolor{red}{"4f78d6f47f5111ef8b42581122bf941e"}?}\\
    \hline
    Scenario:  \textit{List} \newline Task:  \textit{Blur-ID} & \textbf{You're a searcher. You need to output the corresponding list elements based on the instructions and the list below. Please follow the instructions directly without anything else.} \newline List to be retrieved:\newline 1. 4f63efbe7f5111ef8b42581122bf941e\newline 2.Summarize the Rome Statute of the International Criminal Court.\newline  ...\newline 153. 4f7ea64c7f5111ef8b42581122bf941e\newline \textit{Instruction:Randomly select an item from the list above, \textcolor{red}{after} the \textcolor{red}{8th} element.}\\
    \hline
    Scenario:  \textit{List} \newline Task:  \textit{Blur-Element} & \textbf{You're a searcher. You need to output the corresponding list elements based on the instructions and the list below. Please follow the instructions directly without anything else.} \newline List to be retrieved:\newline 1. 4f63efbe7f5111ef8b42581122bf941e\newline 2.Summarize the Rome Statute of the International Criminal Court.\newline  ...\newline 153. 4f7ea64c7f5111ef8b42581122bf941e\newline \textit{Instruction: Retrieve the entry at position \textcolor{red}{8th} in the list. Display it immediately.}\\
    \hline

    Scenario:  \textit{MultiDoc} \newline Task:  \textit{Find-dup-doc} & \textbf{You are a document manager. Here is a collection of documents. Each document includes information such as title, date, source, id, iD2 and specific article content (text). You need to read the documents and follow the instructions to give some information directly, without something else. Also note:\newline 1. Some documents may be missing information such as title or source, which may affect the final output.\newline 2. Some article conent (i.e. values corresponding to the text keyword) may be duplicated.} \newline 
     Documents:\newline ******************** doc-1 ********************\newline text: Following discussions with the PSNI, ... \"\newline source: meeting\newline iD2: bec60ab1-35bc-417d-b378-0b2e86dfcd23\newline id: xWTp7xEyQrqGmGj5p6CdVQ\newline title: 39272756\newline date: 2016-06-20\newline ******************** doc-2 ********************\newline id: SOl-GT2FSyKFYJzBhkYzxw \newline text: What did the participants think about using CD's for backup? ...\newline source: news\newline date: 2001-01-15\newline iD2: 12a0f93d-0f57-46bc-bf28-9ec856fc974a\newline title: tr-sq-167\newline ******************** doc-3 ******************** \newline .... \newline 
    \textit{Instruction: Within the supplied documents, certain documents contain replicated content in their 'text' field, although other fields (such as id, iD2, date, title, source) may be different. Additionally, there could be N sets of documents, each set comprising any number of replicates. Please identify these replicated documents and present \textcolor{red}{iD2} in sequence. The output should have N lines, each line symbolizing a set of replicated documents. Format the output for each document set as depicted in the example. Avoid providing explanations. If a document lacks information in a specific field, use 'None' instead.\newline output example:\newline \textcolor{red}{[["iD2\_1"], ["iD2\_2"], ["iD2\_3"]]\newline[["iD2\_4"], ["iD2\_5"]]}}\\
    \hline

    Scenario:  \textit{MultiDoc} \newline Task:  \textit{Batch-label} & \textbf{You are a document manager. Here is a collection of documents. Each document includes information such as title, date, source, id, iD2 and specific article content (text). You need to read the documents and follow the instructions to give some information directly, without something else. Also note:\newline 1. Some documents may be missing information such as title or source, which may affect the final output.\newline 2. Some article conent (i.e. values corresponding to the text keyword) may be duplicated.} \newline 
     Documents:\newline ******************** doc-1 ********************\newline text: Following discussions with the PSNI, ... \"\newline source: meeting\newline iD2: bec60ab1-35bc-417d-b378-0b2e86dfcd23\newline id: xWTp7xEyQrqGmGj5p6CdVQ\newline title: 39272756\newline date: 2016-06-20\newline ******************** doc-2 ********************\newline id: SOl-GT2FSyKFYJzBhkYzxw \newline text: What did the participants think about using CD's for backup? ...\newline source: news\newline date: 2001-01-15\newline iD2: 12a0f93d-0f57-46bc-bf28-9ec856fc974a\newline title: tr-sq-167\newline ******************** doc-3 ******************** \newline .... \newline 
    \textit{Instruction: Assign labels to documents in order using the provided list of ['11311', '22422', '33233', '44444']. The labeling rules to follow are: \newline \textcolor{red}{ 1. If the document contains both title and source information, mark it as "11311".\newline 2. If the document is missing the source information but not the title, mark it as "22422".\newline 3. If the document is missing title information but not source, mark it as "33233".\newline 4. If the document is missing both title and source information, mark it as "44444".} \newline The tags should be output in JSON dictionary format, for example: \{"doc1":"11311","doc2":"22422"\}}\\
    \hline
    Scenario:  \textit{OneDoc} \newline Task:  \textit{Repeat} & \textbf{There are several different types of KEY SENTENCE in the input text, which are marked by special tags. These special tags a total of six kinds, respectively is <\#Topic\#>, <@argument@>, "<!Transition!>", "<|Summary|>", "<*Evidence*>", "<-Concession->". Different tags represent different types of key sentence. If a sentence in the text is KEY SENTENCE, we will add a special tag with the same attribute to the beginning and end of the sentence. The head tag also contains id order information in the format <type-id>. For example, the head tag with type '\#Topic\#' and id 1 is <\#Topic\#-1>. Also note that when the head tag and tail tag attributes are inconsistent, this means that the sentence is a fake KEY SENTENCE. Please read the input text carefully and give the answer directly according to the instruction requirements.} \newline Input text: Remember the essays you had to write in high school? Topic sentence, introductory paragraph, supporting paragraphs, conclusion.  <\#Topic\#-2>With the result that writing is made to seem boring and pointless.<\#Topic\#> Who cares about symbolism in Dickens? ...\newline \textit{Instruction: Provide \textcolor{red}{4} KEY SENTENCE and their categories directly. Display the output on \textcolor{red}{4} individual lines with each line containing a KEY SENTENCE and its category, separated by \textcolor{red}{|||||}. \newline Output example:\newline [KEY\_SENTENCE\_1] \textcolor{red}{|||||} \#Topic\#\newline [KEY\_SENTENCE\_2] \textcolor{red}{|||||} *Evidence*}\\
    \hline
    Scenario:  \textit{OneDoc} \newline Task:  \textit{Extract} & \textbf{There are several different types of KEY SENTENCE in the input text, which are marked by special tags. These special tags a total of six kinds, respectively is <\#Topic\#>, <@argument@>, "<!Transition!>", "<|Summary|>", "<*Evidence*>", "<-Concession->". Different tags represent different types of key sentence. If a sentence in the text is KEY SENTENCE, we will add a special tag with the same attribute to the beginning and end of the sentence. The head tag also contains id order information in the format <type-id>. For example, the head tag with type '\#Topic\#' and id 1 is <\#Topic\#-1>. Also note that when the head tag and tail tag attributes are inconsistent, this means that the sentence is a fake KEY SENTENCE. Please read the input text carefully and give the answer directly according to the instruction requirements.} \newline Input text: Remember the essays you had to write in high school? Topic sentence, introductory paragraph, supporting paragraphs, conclusion.  <\#Topic\#-2>With the result that writing is made to seem boring and pointless.<\#Topic\#> Who cares about symbolism in Dickens? ...\newline \textit{Instruction: Gather every instance of KEY SENTENCE classified as \textcolor{red}{\#Topic\#}. The output should be a Json list arranged by ids. If none are found, provide an empty array. \newline Output Example 1: [KEY SENTENCE1, KEY SENTENCE2, KEY SENTENCE3,...]\newline Output Example 2: []}\\
    \hline
    Scenario:  \textit{OneDoc} \newline Task:  \textit{QA} & \textbf{There are several different types of KEY SENTENCE in the input text, which are marked by special tags. These special tags a total of six kinds, respectively is <\#Topic\#>, <@argument@>, "<!Transition!>", "<|Summary|>", "<*Evidence*>", "<-Concession->". Different tags represent different types of key sentence. If a sentence in the text is KEY SENTENCE, we will add a special tag with the same attribute to the beginning and end of the sentence. The head tag also contains id order information in the format <type-id>. For example, the head tag with type '\#Topic\#' and id 1 is <\#Topic\#-1>. Also note that when the head tag and tail tag attributes are inconsistent, this means that the sentence is a fake KEY SENTENCE. Please read the input text carefully and give the answer directly according to the instruction requirements.} \newline Input text: Remember the essays you had to write in high school? Topic sentence, introductory paragraph, supporting paragraphs, conclusion.  <\#Topic\#-2>With the result that writing is made to seem boring and pointless.<\#Topic\#> Who cares about symbolism in Dickens? ...\newline \textit{{Instruction: Is \textcolor{red}{"With the result that writing is made to seem boring and pointless."} the KEY SENTENCE of the text? Indicate \textcolor{red}{"True"} for yes, \textcolor{red}{"False"} for no.}}\\
    \hline
  \caption{\label{tab:allprompts}
    Examples of all tasks in \ours. 
    \textbf{Bold} denotes the scenario description $D$; normal denotes the context $X$; \textit{italics} denotes instruction $I$, with \textcolor{red}{red} indicate the instruction variables $var$, and the remaining black parts correspond to the instruction template $tpl$.
  }
\end{longtable}

\end{document}